\documentclass[final, 5p]{elsarticle}

\usepackage{lineno,hyperref}
\usepackage{amsmath,graphicx}
\usepackage{bm}
\usepackage{subcaption}
\usepackage[bb=boondox]{mathalfa}
\usepackage{color}
\journal{Neurocomputing}

\begin{document}

\begin{frontmatter}

\title{Pose Guided Structured Region Ensemble Network for Cascaded Hand Pose Estimation}

\author[mymainaddress]{Xinghao Chen}

\author[mymainaddress]{Guijin Wang\corref{mycorrespondingauthor}}
\cortext[mycorrespondingauthor]{Corresponding author: wangguijin@tsinghua.edu.cn (G. Wang)}

\author[secondaddress]{Hengkai Guo}

\author[mymainaddress]{Cairong Zhang}

\tnotetext[note]{Email addresses: chen-xh13@mails.tsinghua.edu.cn (X. Chen), guohengkaighk@gmail.com (H. Guo), zcr17@mails.tsinghua.edu.cn (C. Zhang). Work was done when H. Guo was with Dept. of EE, Tsinghua University.}

\address[mymainaddress]{Department of Electronic Engineering, Tsinghua University, Beijing 100084, China}
\address[secondaddress]{AI Lab, Bytedance Inc., Beijing, China}

\begin{abstract}
Hand pose estimation from single depth images is an essential topic in computer vision and human computer interaction. Despite recent advancements in this area promoted by convolutional neural networks, accurate hand pose estimation is still a challenging problem. In this paper we propose a novel approach named as Pose guided structured Region Ensemble Network (Pose-REN) to boost the performance of hand pose estimation. Under the guidance of an initially estimated pose, the proposed method extracts regions from the feature maps of convolutional neural network and generates more optimal and representative features for hand pose estimation. The extracted feature regions are then integrated hierarchically according to the topology of hand joints by tree-structured fully connections to regress the refined hand pose.
The final hand pose is obtained by an iterative cascaded method.
Comprehensive experiments on public hand pose datasets demonstrate that our proposed method outperforms state-of-the-art algorithms.
\end{abstract}

\begin{keyword}
Hand Pose Estimation\sep Convolutional Neural Network\sep Human Computer Interaction \sep Depth Images
\end{keyword}

\end{frontmatter}


\section{Introduction}
\label{sec:intro}

Accurate 3D hand pose estimation is one of the most important techniques in human computer interaction and virtual reality~\cite{erol2007vision}, since it can provide fundamental information for interacting with objects and performing gestures~\cite{chen2017motion, de2016skeleton}. Hand pose estimation from single depth images has attracted broad research interests in recent years~\cite{supancic2015depth, tompson2014real, tang2015opening, YeSpatialHandECCV2016, Xu2017, guo2017region, wan2017crossing, ge2017threedcnn} thanks to the availability of depth cameras~\cite{zhang2012microsoft, wang2013depth, shi2015high, keselman2017intel}, such as Microsoft Kinect, Intel Realsense Camera etc. However, hand pose estimation is an extremely challenging problem due to the severe self-occlusion, high complexity of hand articulation, noises and holes in depth image, large variation of viewpoints and self-similarity of fingers etc.

\begin{figure*}[!htb]
  \centering
    \centerline{\includegraphics[width=0.9\linewidth]{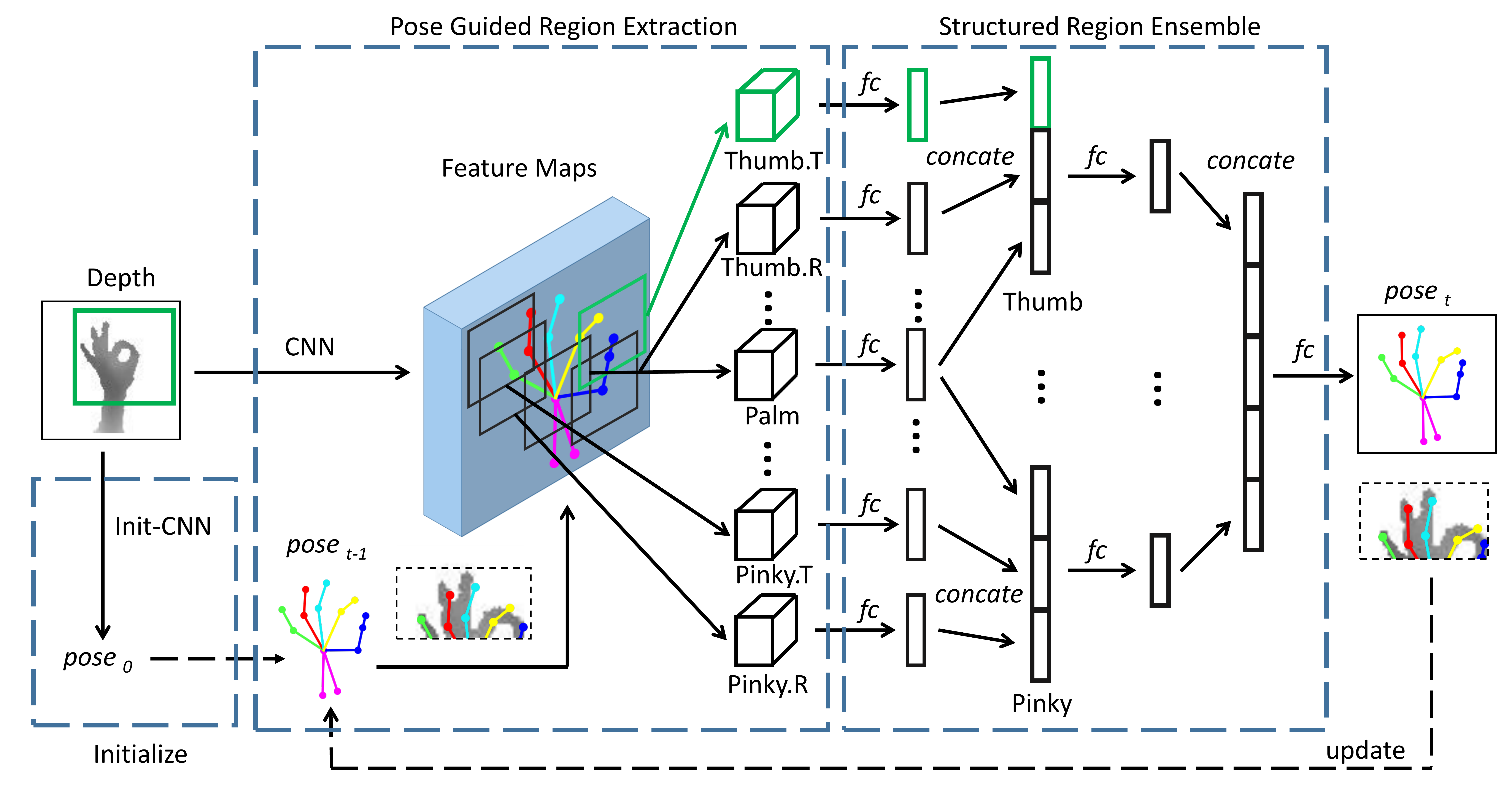}}
    \caption{The framework of our proposed pose guided structured region ensemble network (Pose-REN). A simple CNN (Init-CNN) predicts $pose_0$ as the initialization of the cascaded framework. Feature regions are extracted from the feature maps generated by a CNN under the guidance of $pose_{t-1}$ and hierarchically fused using a tree-like structure. $pose_t$ is the refined hand pose obtained by our proposed Pose-REN and will be used as the guidance in next stage.}
\label{fig:framework}
\end{figure*}

Hand pose estimation has achieved great advancements by convolutional neural networks (CNNs).
CNN-based data-driven methods either predict heatmaps of hand joints~\cite{tompson2014real, ge2016robust} and infer hand pose from heatmaps, or directly regress the 3D coordinates of hand joints~\cite{oberweger2015hands, oberweger2015training, YeSpatialHandECCV2016, madadi2017end, wan2017crossing, zhou2016model}. In either ways, features are critical for the performance of hand pose estimation. Prior works mainly focused on incorporating prior knowledge into CNN~\cite{oberweger2015hands, zhou2016model} or using error feedback~\cite{oberweger2015training} and spatial attention design~\cite{YeSpatialHandECCV2016}. However, few of prior works have paid attentions to extracting more optimal and representative features of CNN.
Ye et al.~\cite{YeSpatialHandECCV2016} used spatial attention module to select and transform features to a canonical space.
Guo et al.~\cite{guo2017region, wang2018region} proposed the region ensemble network (REN) that divides the feature maps of last convolutional layer into several spatial regions and integrates them in fully connected layers. All aforementioned works haven't fully exploit optimal features of CNN for hand pose estimation.

In this paper, we propose a novel method called pose guided structured region ensemble network (Pose-REN) to boost the performance of hand pose estimation, as shown in Figure~\ref{fig:framework}. Upon an iterative refinement procedure, our proposed method takes a previously estimated pose as input and predicts a more accurate result in each iteration. We present a novel feature extraction method under the guidance of previous predicted hand pose to get optimal and representative features for hand pose estimation.
Furthermore, inspired by hierarchical recurrent neural network~\cite{du2015hierarchical}, we present a hierarchical method to fuse features of different joints according to the topology of hand. Features from joints that belong to the same finger are integrated in the first layer and features from all fingers are fused in the following layers to predict the final hand pose.

We evaluate our proposed method on three public hand pose benchmarks~\cite{tang2014latent, tompson2014real, sun2015cascaded}. Compared with state-of-the-art methods, our method has achieved the best performance. Extensive ablation analyses illustrate the contributions of different components of the framework and robustness of our proposed method.

The remainder of this paper is organized as follows. In Section~\ref{sec:related-work}, we review prior works that are highly related to our proposed method. In Section~\ref{sec:method}, we present details about our proposed pose guided structured region ensemble network. Evaluations on public datasets and ablation studies are provided in Section~\ref{sec:exp}. Section~\ref{sec:conclusion} gives a brief conclusion of this paper.

\section{Related Work}
\label{sec:related-work}
In this section we briefly review related works of our proposed method. Firstly we will review recent algorithms for depth based hand pose estimation. Since our method basically builds upon cascaded framework, we will introduce the cascaded methods for hand pose estimation. Finally, we will review related works about the hierarchical structure of neural network, as the hierarchical structured connections are utilized in our method.

\subsection{Depth-based Hand Pose Estimation}
Recent approaches of hand pose estimation are generally categorized into three classes: discriminative methods~\cite{tompson2014real, tang2014latent, oberweger2015hands, sun2015cascaded, wan2016hand, wan2017crossing, ge2016robust, valentin2016learning, zhang2016learning, ge2017threedcnn, guo2017region}, generative methods~\cite{tagliasacchi2015robust, tkach2016sphere, joseph2016fits, taylor2016efficient} and hybrid methods~\cite{krejov2015combining, tang2015opening, choi2015collaborative, sridhar2015fast, sharp2015accurate, YeSpatialHandECCV2016, zhou2016model}.
Comprehensive review and analysis on depth based 3D hand pose estimation can be found in~\cite{supancic2015depth}.

Generative methods fit a predefined hand model to the input data using optimization algorithms to obtain the optimized hand pose, such as PSO (particle swarm optimization)~\cite{sharp2015accurate}, ICP (Iterative Closest Point)~\cite{tagliasacchi2015robust} and their combination (PSO-ICP)~\cite{qian2014realtime}. Hand-crafted energy functions that describe the distance between the hand model and input image are utilized in prior works, such as golden energy~\cite{sharp2015accurate} and silver energy~\cite{tang2015opening}. Several kinds of hand model have been adopted, including sphere model~\cite{qian2014realtime}, sphere-meshes model~\cite{tkach2016sphere}, cylinder model~\cite{tagliasacchi2015robust} and mesh model~\cite{sharp2015accurate}.
Generative methods are robust for self-occlusive areas or missing areas and ensure to output plausible hand pose. However, they need a complex and time-consuming optimizing procedure and are likely to trap into local optimizations.

Discriminative methods directly learn a predictor from the labelled training data. The predictor either predicts the probability maps (heatmaps) of each hand joints~\cite{tompson2014real, ge2016robust} or directly predicts the 3D hand joint coordinates~\cite{oberweger2015hands, guo2017region}. The most frequently used methods for predictor are random forest~\cite{tang2013real, tang2014latent, liang2014parsing, tang2015opening, sun2015cascaded} and convolutional neural network~\cite{tompson2014real, oberweger2015hands, guo2017region, wan2017crossing, ge2017threedcnn}.
Discriminative methods do not require any complex hand model and are totally data-driven, which are fast and appropriate for real-time applications. Guo et al.~\cite{guo2017region, wang2018region} proposed a region ensemble network (REN) that greatly promoted the performance of hand pose estimation based on a single network. Region ensemble network divides the feature maps of last convolutional layer into several spatial regions and integrates them in fully connected layers. However, REN extracts the feature regions using a uniform grid and all features are treated equally, which is not optimal to fully incorporate the spatial information of feature maps and obtain highly representative features.

Hybrid methods try to combine the discriminative and generative methods to achieve better hand pose estimation performance. Some works adopted the generative methods after obtaining initial results by discriminative methods~\cite{krejov2015combining, sridhar2015fast, sharp2015accurate}. Zhou et al.~\cite{zhou2016model} proposed to incorporate a hand model into the CNN, which exploits the constraints of the hand and ensures the geometric validity of the estimated pose. However, hybrid methods have to predefine the properties of the hand model, such as the length of bones. Oberweger et al.~\cite{oberweger2015training} proposed a data-driven hybrid method, which learns to generate a depth image from hand pose. However, the generation of depth images is likely affected by the errors of annotations.

Our proposed method basically falls into the category of discriminative method and does not rely on any predefined hand model. Compared with prior CNN-based discriminative methods, our proposed method directly predicts the 3D locations of hand pose using a cascaded framework without any postprocessing procedure. What's more, our proposed pose guided structured region ensemble network (Pose-REN) can learn better features for hand pose estimation by incorporating guided information of previously estimated hand pose into the feature maps and improve the performance of our method.

Although our proposed Pose-REN follows the idea of feature region ensemble as REN~\cite{guo2017region}, there are several essential differences between Pose-REN and REN~\cite{guo2017region}: 1) Different from REN that uses grid region feature extraction, the proposed Pose-REN fully exploits an initially estimated hand pose as the guided information to extract more representative features from CNN, which is shown to have a large impact for hand pose estimation problem, as discussed in Section~\ref{sec:ablation-pose-guided}. 2) Instead of simple feature fusion as adopted in REN, our Pose-REN presents a structured region ensemble strategy that better models the connections and constraints between different joints in the hand. 3) The Pose-REN is a common framework that can easily be compatible with any existing methods (for example, Feedback~\cite{oberweger2015training}, DeepModel~\cite{zhou2016model} etc.) by using them to produce initial estimations for Pose-REN.

\subsection{Cascaded Method}
The cascaded framework has been widely used in face alignment~\cite{zhu2015face, chen2014joint, kowalski2017deep}, human pose estimation~\cite{toshev2014deeppose, carreira2016human} and has also shown good performances in the problem of hand pose estimation~\cite{sun2015cascaded, oberweger2015training, YeSpatialHandECCV2016}.

Sun et al.~\cite{sun2015cascaded} proposed a method to iteratively refine the hand pose using hand-crafted 3D pose index features that are invariant to viewpoint transformation. Oberweger et al.~\cite{oberweger2015hands} proposed a post-refinement method to refine each joint independently using multiscale input regions centered on the initially estimated hand joints. These works have to train multi models for refinement and independently predict different parts of hand joints while our proposed needs only one model to iteratively improve the estimated hand pose.

Oberweger et al.~\cite{oberweger2015training} presented a feedback loop framework for hand pose estimation. One discriminative network is used to produce initial hand pose. A depth image is then generated from the initial hand pose using a generative CNN and an updater network improves the hand pose by comparing the synthetic depth image and input depth image. However, the depth synthetic network is highly sensitive to the annotation errors of hand poses.

Ye et al.~\cite{YeSpatialHandECCV2016} integrated cascaded and hierarchical regression into a CNN framework using spatial attention mechanism. The partial hand joints are iteratively refined using transformed features generated by spatial attention module. In their method, the features in cascaded framework are generated by a initial CNN and remain unchanged in each refinement stage except for the spatial transformation. In our proposed method, feature maps are updated in each cascaded stage using an end-to-end framework, which will help to learn more effective features for hand pose estimation.

Our Pose-REN also adopts the cascaded framework. Different from the above prior methods, we present a novel feature extraction method under the guidance of previous predicted hand pose to get optimal and representative features from CNN. What's more, Pose-REN explicitly models the constraints and relations between different hand joints using structured region ensemble strategy, which is a novel method to improve the robustness and performance of hand pose estimation.

\subsection{Hierarchical Structure of Neural Network}

Du et al.~\cite{du2015hierarchical} proposed a hierarchical recurrent neural network (RNN) for skeleton-based human action recognition. The whole skeleton is divided into five parts and fed into different branches of the RNN. Different parts of skeleton are hierarchically fused to generated higher-level representations.
Madadi et al.~\cite{madadi2017end} proposed a tree-shape structure of CNN which regresses local poses at different branches and fuses all features in the last layer. In their structure, features of different partial poses are learned independently except for sharing features in very early layers. In contrast, our method shares features in the convolutional layers for all joints and hierarchically fuses different regions from feature maps to finally estimate the hand pose. The shared  features enables better representation of hand pose and the hierarchical structure of feature fusion can better model the correlation of different hand joints.

\section{Pose Guided Structured Region Ensemble Network}
\label{sec:method}
In this section, we first give an overview of Pose-REN in Section~\ref{sec:method-overview}. After that we will provide detailed elaboration about extracting regions from the feature maps under the guidance of a hand pose in Section~\ref{sec:pren}. In Section~\ref{sec:method-structure} we present the details of fusing feature regions using hierarchically structured connection. Finally, the training strategy and implementation details are given in Section~\ref{sec:training} and Section~\ref{sec:implementation}.

\subsection{Overview}
\label{sec:method-overview}
The framework of our proposed method is depicted in Figure~\ref{fig:framework}. A simple CNN (denoted as Init-CNN) predicts an initial hand pose $pose_0$, which is used as the initialization of the cascaded framework. The proposed framework takes a previously estimated hand pose $pose_{t-1}$ and the depth image as input. The depth image is fed into a CNN to generate feature maps. Feature regions are extracted from these feature maps under the guidance of the input hand pose $pose_{t-1}$. The insight of our proposed method is that features around the location of a joint contribute more while other features like corner regions are less important. Afterwards, features from different joints are hierarchically integrated using the structured connection to regress the refined hand pose $pose_t$. The images in dash rectangles show the close-up results of $pose_{t-1}$ and $pose_{t}$. It can be seen that the network refines the hand pose gradually.

Our method aims to estimate the 3D hand pose from a single depth image in a cascaded framework. Specifically, given a depth image $\mathcal D$, the 3D locations $\mathcal P=\{p_i = (p_{xi}, p_{yi}, p_{zi})\}_{i=1}^{J}$ of $J$ hand joints are inferred.  Given a previously estimated hand pose result ${\mathcal P}^{t-1}$ in stage $t-1$, our method uses the learned regression model ${\mathcal R}$ to refine the hand pose in stage $t$.
\begin{equation}\label{eq:cascaded}
  {\mathcal P}^t = {{\mathcal R}}({\mathcal P}^{t-1}, {\mathcal D})
\end{equation}
After $T$ stages, we get the final estimated hand pose ${\mathcal P}^T$ for the input depth image $\mathcal D$.
\begin{equation}\label{eq:cascaded-final}
  {\mathcal P}^T = {{\mathcal R}}({\mathcal P}^{T-1}, {\mathcal D})
\end{equation}
It should be noted that only one same model ${\mathcal R}$ is used in every stage of refinement in the inference phase, see Section~\ref{sec:training} for details.

\subsection{Pose Guided Region Extraction}
\label{sec:pren}

\begin{figure}[!tb]
  \centering
    \centerline{\includegraphics[width=\linewidth]{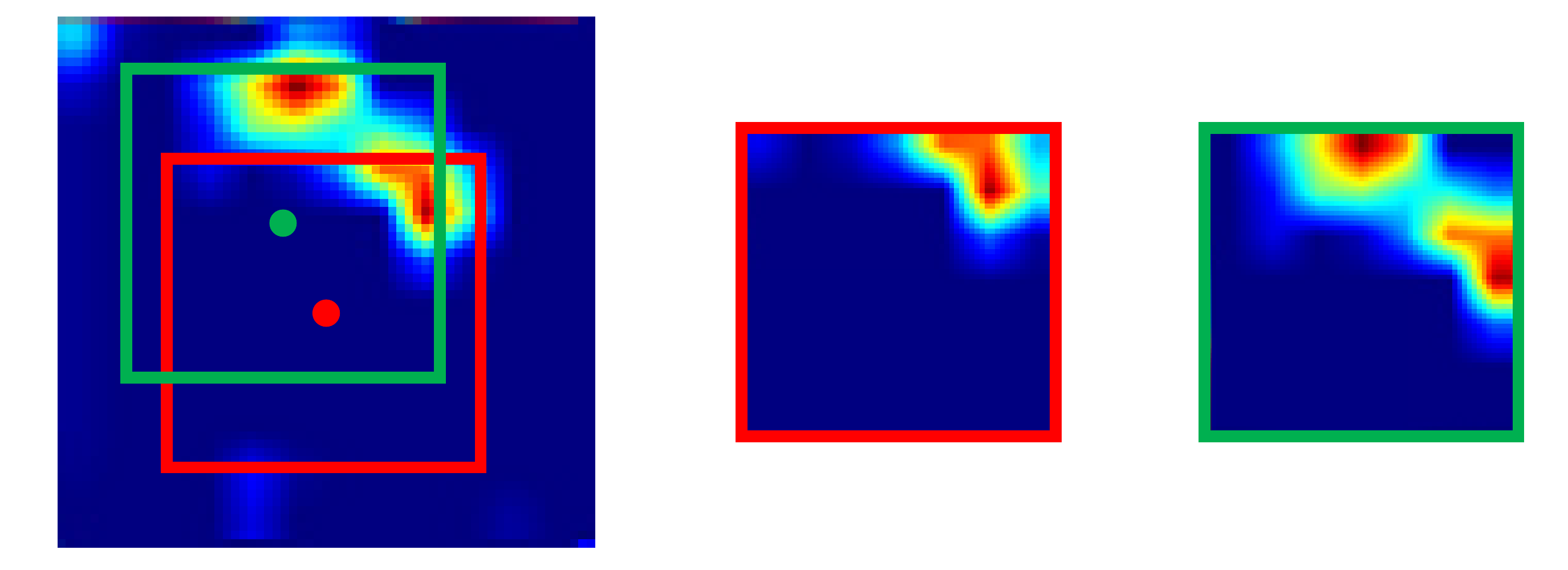}}
  \caption{The scheme of our proposed pose guided region extraction. The green and red dots represent two hand joints from previously estimated hand pose. The rectangles of different colors are the corresponding feature regions extracted from the feature maps.}
\label{fig:pose-feature}
\end{figure}

We first use a standard convolutional neural network (CNN) with residual connections to generate feature maps. The backbone architecture of CNN for generating feature maps used in our method is the same as the baseline network in~\cite{guo2017region}, with $6$ convolutional layers and $2$ residual connections. Each convolutional layer is followed by a Rectified Linear Unit (ReLU)~\cite{maas2013rectifier} as the activation function and every $2$ convolutional layers are followed by a max pooling layer. The residual connections are added between max pooling layers.

Denote feature maps from the last convolutional layer as $\mathcal F$ and the estimated hand pose from previous stage as $\mathcal P^{t-1}=\{(p_{xi}^{t-1}, p_{yi}^{t-1}, p_{zi}^{t-1})\}_{i=1}^J$. We use $\mathcal P^{t-1}$ as the guidance to extract feature regions from $\mathcal F$. Specifically, for the $i^{th}$ hand joint, We first project the real-world coordinates into the image pixel coordinates using the intrinsic parameters of the depth camera, as shown in Eq.~\ref{eq:xyz2uvd}.
\begin{equation}\label{eq:xyz2uvd}
  (p_{ui}^{t-1}, p_{vi}^{t-1}, p_{di}^{t-1}) = \mathnormal {proj}(p_{xi}^{t-1}, p_{yi}^{t-1}, p_{zi}^{t-1})
\end{equation}

The feature region for this joint is then cropped using a rectangular window which can be defined by a tuple $(b_{ui}^{t}, b_{vi}^{t}, w, h)$, where $b_{ui}^{t}$ and $b_{vi}^{t}$ is the coordinates of top-left corner, $w$ and $h$ is the width and height of the cropped feature region.
The coordinates of the rectangular window are calculated by normalizing and converting the original coordinates $(p_{ui}^{t-1}, p_{vi}^{t-1}, p_{di}^{t-1})$ into coordinates in feature maps.

The extracted feature region for hand joint $i$ is then obtained by cropping the feature maps within the rectangular window:
\begin{equation}\label{eq:region_i}
  \mathcal F^t_i = \mathnormal {crop}(\mathcal F; b_{ui}^{t}, b_{vi}^{t}, w, h)
\end{equation}
where the function $\mathnormal {crop(\mathcal F; b_{u}, b_{v}, w, h)}$ means extracting the region specified by a rectangular window $(b_{u}, b_{v}, w, h)$ from $\mathcal F$.

Figure~\ref{fig:pose-feature} gives an example of pose guided region extraction. The left image is a feature map from the last convolutional layer of the CNN. It should be noted the feature maps usually contains multiple channels, we only use one channel of them to depict how to crop a region guided by a joint. The green dot and red dot indicate two joints (palm center joint and Metacarpophalangeal joint for middle finger respectively) from the previously estimated hand pose. The green and red rectangles are the corresponding cropped windows. The images in middle and right columns show the extracted feature regions for these two joints.

\subsection{Structured Region Ensemble}
\label{sec:method-structure}
\begin{figure}[tb]
  \centering
    \centerline{\includegraphics[width=\linewidth]{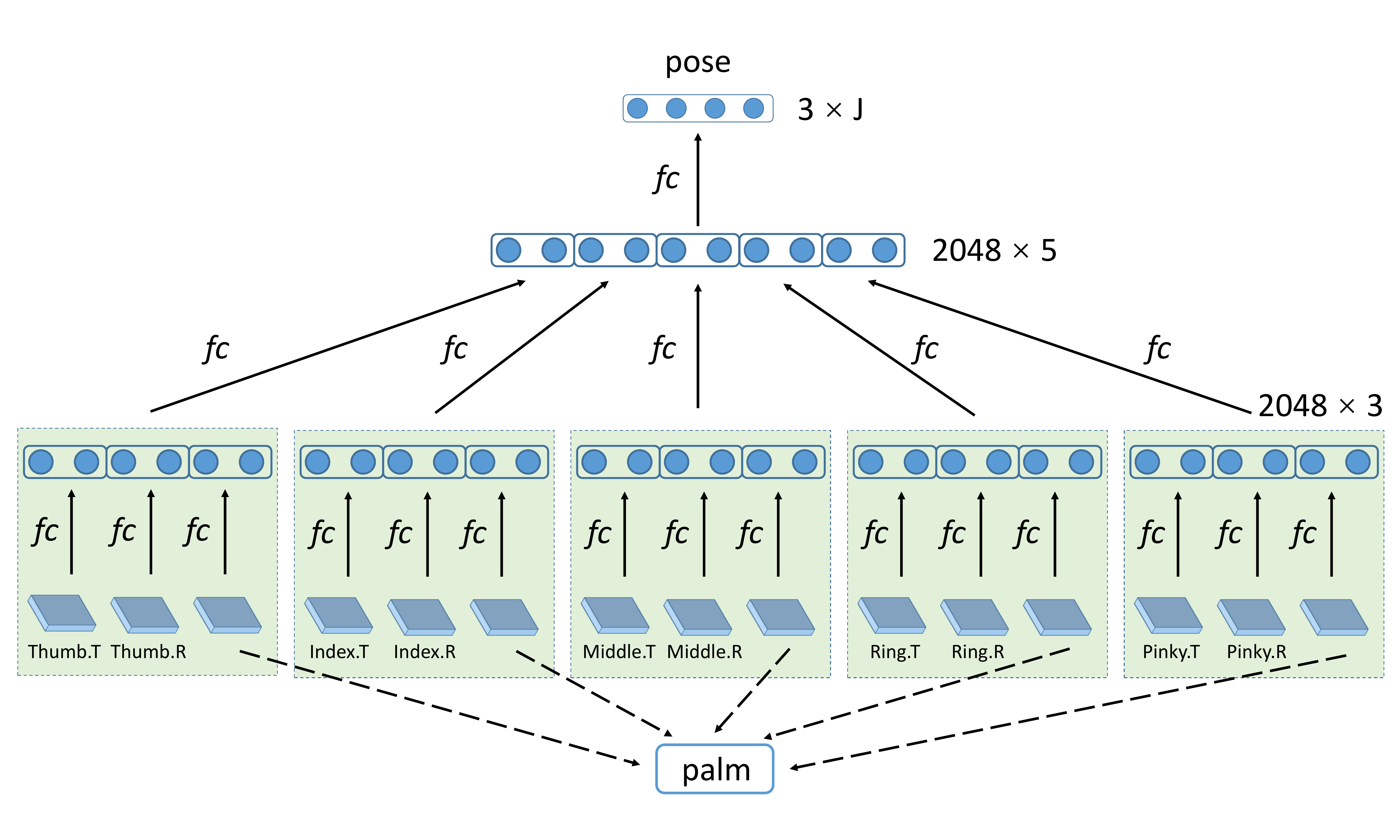}}
  \caption{The architecture of the proposed structured region ensemble method. Features from the joints of the same finger (including the palm joint) are fused first. Afterwards, features of different fingers are fused to regress the final hand pose.}
\label{fig:structure-region-ensemble}
\end{figure}

In the previous section we have described how to extract feature regions from the feature maps for each joint using the guidance of previously estimated hand pose. One intuitional way to fuse these feature regions is to connect each region with fully connected ($fc$) layers respectively and then fuse these layers to regress the final hand pose, which is adopted in REN~\cite{guo2017region}.

Human hand is a highly complex articulated object. Therefore, there are many constraints and correlations between different joints~\cite{lin2000modeling, wu2001hand}. Independently connecting feature regions with $fc$ layers and fusing them in the last layer can not fully adopt these constraints. Inspired by hierarchical recurrent neural network~\cite{du2015hierarchical}, in this paper we adopt hierarchically structured region ensemble strategy to better model the constraints of hand joints, as shown in Figure~\ref{fig:structure-region-ensemble}. First a set of feature regions $\{\mathcal F^t_j\}_{j=1}^{M}$ are fed into $fc$ layers respectively.
\begin{equation}\label{eq:fc1}
  h_j^{l_1} = \mathnormal {fc} (\mathcal F^t_j),\quad j = 1, \ldots, M
\end{equation}
Where $M$ is the number of regions extracted from the feature maps.

Next, $\{h_j^{l_1}\}_{j=1}^{M}$ are integrated hierarchically according the topology structure of hand. Specifically, denote the indices of joints that belong to the $i^{th}$ finger as $\{I_j^i\}_{j=1}^{M_i}$, where $M_i$ is the number of joints that belong to the $i^{th}$ finger. All joints that belong to the same finger are concatenated (denote as $concate$) and then fed into a $fc$ layer, as shown in Eq.~\ref{eq:concate1} and Eq.~\ref{eq:fc2}.
\begin{equation}\label{eq:concate1}
  \bar{h}_i^{l_1} = \mathnormal {concate} (\{h_{I^i_j}^{l_1}\}_{j=1}^{M_i}),\quad i = 1, \ldots, 5
\end{equation}
\begin{equation}\label{eq:fc2}
  h_i^{l_2} = \mathnormal {fc} (\bar{h}_i^{l_1}),\quad i = 1, \ldots, 5
\end{equation}
Afterwards, features from different fingers $\{h_i^{l_2}\}_{i=1}^5$ are concatenated and fed into a $fc$ layer to regress the final hand pose $\mathcal P^t \in \mathbb R^{3 \times J}$.
\begin{equation}\label{eq:concate2}
  \bar{h}^{l_2} = \mathnormal {concate} (\{h_i^{l_2}\}_{i=1}^5)
\end{equation}
\begin{equation}\label{eq:fc3}
  \mathcal P^t = \mathnormal {fc} (\bar{h}^{l_2})
\end{equation}
Each $fc$ layer in Eq.~\ref{eq:fc1} and Eq.~\ref{eq:fc2} has a dimension of $2048$ nodes. They are followed by ReLU layers and dropout layers with dropout rate of $0.5$. The last $fc$ layer output a $3 \times J$ vector $\mathcal P^t$ which represents the 3D locations of hand pose.

\subsection{Training}
\label{sec:training}
Denote the original training set as
\begin{equation}
\mathcal T^0 = \{(\mathcal D_i, \mathcal P^0_i, \mathcal P^{gt}_i)\}_{i=1}^{N_{\mathcal T}}
\end{equation}
where $N_{\mathcal T}$ is the number of training samples, $\mathcal D_i$ is the depth image, $\mathcal P^0_i$ is the initially estimated hand pose and $\mathcal P^{gt}_i$ is the corresponding ground truth of hand pose.

In stage $t$, a regression model $\mathcal R^t$ is trained using $\mathcal T^{t-1}$. Using this model, we can obtain the refined hand pose for each sample in training set.
\begin{equation}\label{eq:refine_trian}
  {\mathcal P}^t_i = {{\mathcal R^t}}({\mathcal P}_i^{t-1}, {\mathcal D})
\end{equation}
we add the refined samples $\overline{\mathcal T^t} = \{(\mathcal D_i, \mathcal P^{t}_i, \mathcal P^{gt}_i)\}_{i=1}^{N_{\mathcal T}}$ to the training set, generating an augmented training set $\mathcal T^t$.
\begin{equation}\label{eq:union_trian}
\mathcal T^t = \mathcal T^{t-1} \bigcup \overline{\mathcal T^t}
\end{equation}
Again, we train a model $\mathcal R^{t+1}$ in stage $t+1$ using $\mathcal T^{t}$ and iteratively repeat this process until reaching the maximum iteration $T$. The trained model $\mathcal R^T$ is the final model used in the inference phase to refine the initial hand pose iteratively, as described in Eq.~\ref{eq:cascaded} and Eq.~\ref{eq:cascaded-final}.

\subsection{Implementation Details}
\label{sec:implementation}
We implemented our proposed method using Caffe~\cite{jia2014caffe}. RoI Pooling layer~\cite{girshick2015fast} was used to facilitate the implementation of pose guided region extraction.

We used the baseline network in~\cite{guo2017region} as the Init-CNN to produce initial poses for our method. Generally speaking, any existing hand pose estimation algorithms can be adopted as the initialization method of Pose-REN. We will further discuss the effect of different initializations in Section~\ref{sec:ablation-init}, including generalization of our pre-trained model to other initializations in inference phase and robustness of Pose-REN to other initializations.

\paragraph{\textbf{Preprocessing}} Similar to previous methods~\cite{oberweger2015training, guo2017region}, we extracted a fix-sized cube from the input depth image. The center of the cube was determined by calculating the centroid of mass of the hand region. The extracted cube was then resized into a patch with size of $96 \times 96$ and the depth values within it were normalized into $[-1, 1]$. Besides, depth values that were outside the cube were truncated according to the size of cube, providing robustness to invalid depth values. The idea of extracting a fix-sized cube is to ensure invariance of the hand size to the distance to the camera.

\paragraph{\textbf{Training}} We first trained the Init-CNN to obtain initial hand pose. After that, we used the weights of trained Init-CNN to initialize Pose-REN and train the network. The whole network was trained using stochastic gradient descent (SGD) with a batch size of $128$ and a momentum of $0.9$. A weight decay of $0.0005$ was also adopted for the network. The learning rate was set to $0.001$ and divided by $10$ after every $25$ epochs. The model was trained for $100$ epochs for each stage and totally trained for two stages.

We followed several good practices that have been proved to be quite effective for hand pose estimation~\cite{guo2017region}, including random data augmentation, smooth $L_1$ loss. For data augmentation, we applied random scaling of $[0.9, 1.1]$, random translation of $[-10, 10]$ pixels and random rotation of $[-180, 180]$ degrees to the depth image. We used smooth $L_1$ loss to achieve less sensitivity to the outliers.

\begin{figure}[!tb]
  \centering
    \centerline{\includegraphics[width=\linewidth]{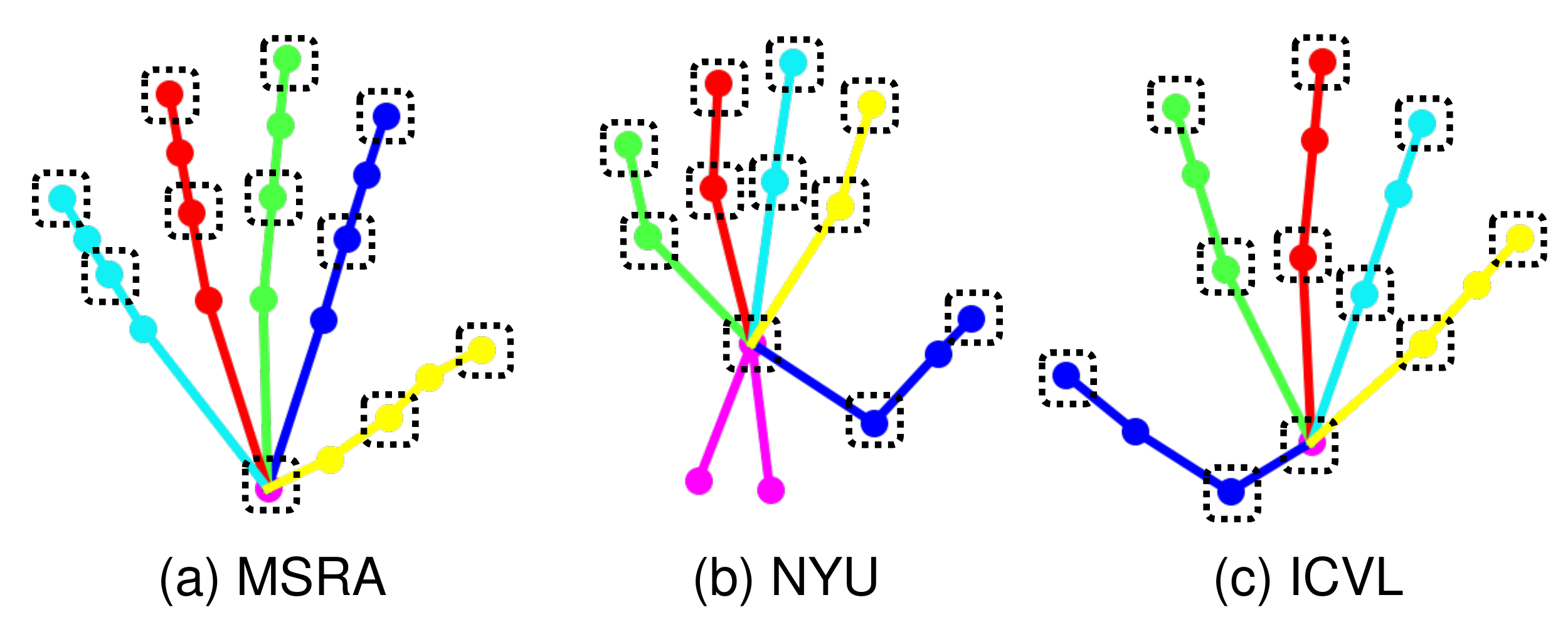}}
  \caption{The subset of joints used in pose guided region extraction. The joints circled by dash rectangles are used when extracting feature regions under the guidance of previous joints. Totally $M=11$ joints are used, including a joint for the palm, two joints for the root and tip of each finger.}
\label{fig:skeleton-setting}
\end{figure}

\paragraph{\textbf{Parameter settings}} Different datasets have the different number of hand joints, e.g. $21$ joints in MSRA dataset and $16$ joints in ICVL dataset. To balance the complexity of model and accuracy, we only used part of joints as the guidance to extract feature regions. Specifically, $11$ out of all joints were used, as shown in Figure~\ref{fig:skeleton-setting}. The joints circled by dash rectangles were used, with $M_i=3$ for each finger, including a joint for the palm, a joint for the root of finger and a joint the tip of finger. It should be noted that despite part of joints ($M=11$) are utilized as the guidance to extract features, the network still predicts the locations of all joints. The insights behind are that the overlaps of different feature regions make sure the covering of almost all important features even only a part of the joints is used.

In our experiments, the size of extracted region was set to $(w,h) = (7,7)$. In inference phase, the number of iterations was set to $T=3$, which will be further discussed in Section~\ref{sec:ablation-iter}.

\begin{figure*}[!tb]
  \centering
  \begin{subfigure}[t]{0.5\textwidth}
    \centering
    \centerline{\includegraphics[width=\linewidth]{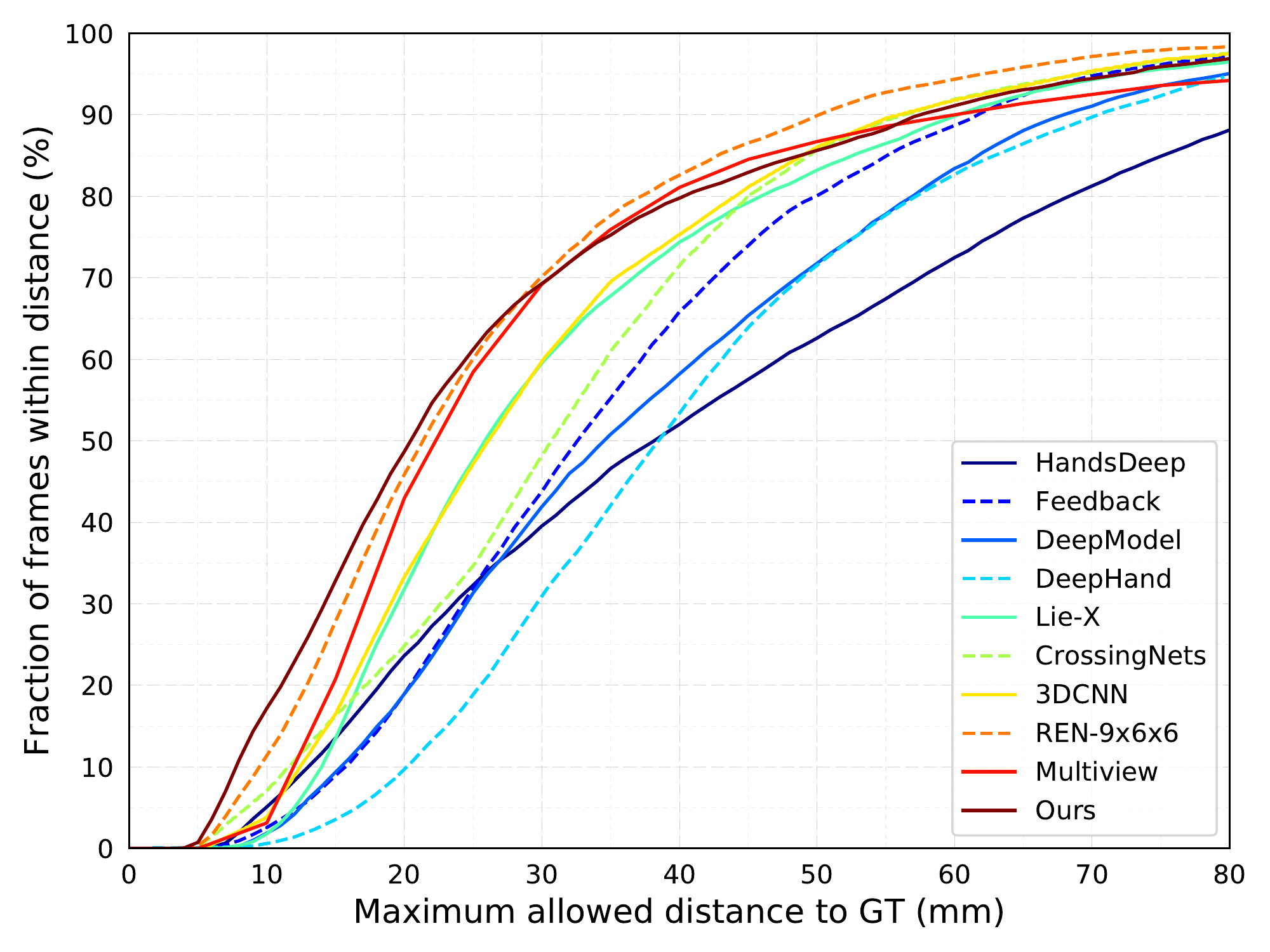}}
  \end{subfigure}%
  \begin{subfigure}[t]{0.5\textwidth}
    \centering
    \centerline{\includegraphics[width=\linewidth]{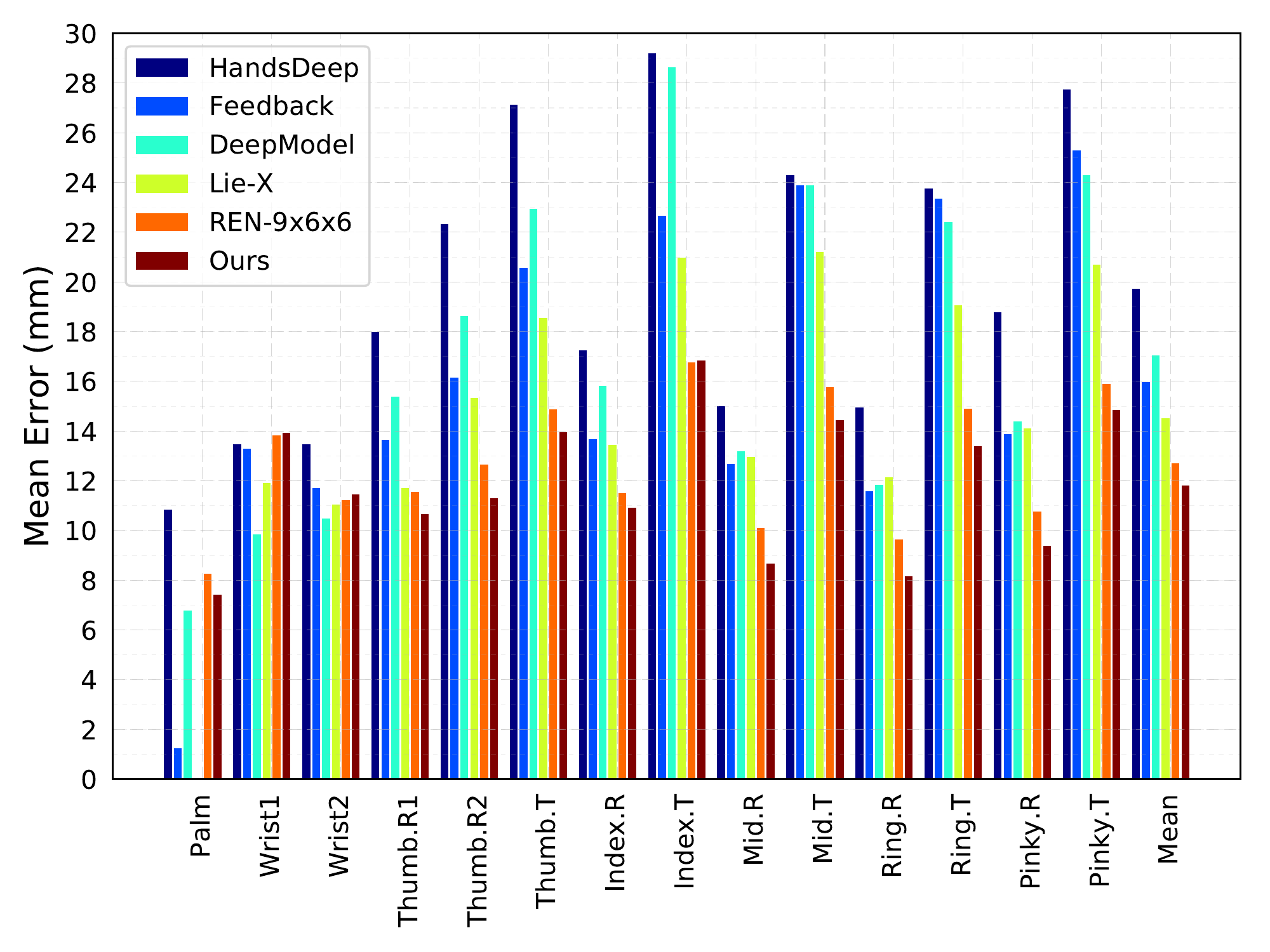}}
  \end{subfigure}%
\caption{Comparison of our approach with state-of-the-art methods on NYU dataset. Left: the proportion of good frames
over different error thresholds. Right: per-joint errors.}
\label{fig:nyu-results}
\end{figure*}

\begin{figure*}[!tb]
  \centering
  \begin{subfigure}[t]{0.5\textwidth}
    \centering
    \centerline{\includegraphics[width=\linewidth]{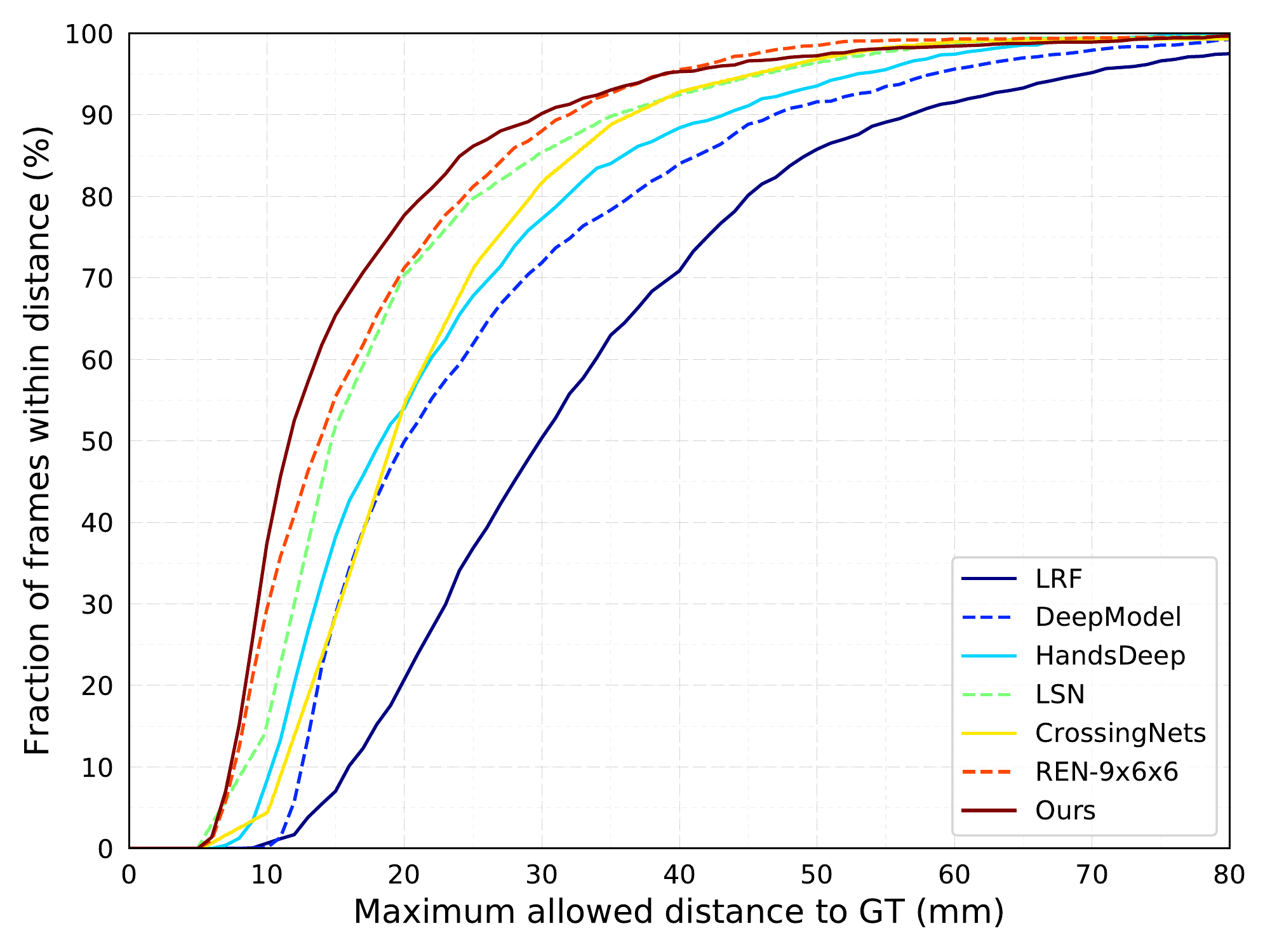}}
  \end{subfigure}%
  \begin{subfigure}[t]{0.5\textwidth}
    \centering
    \centerline{\includegraphics[width=\linewidth]{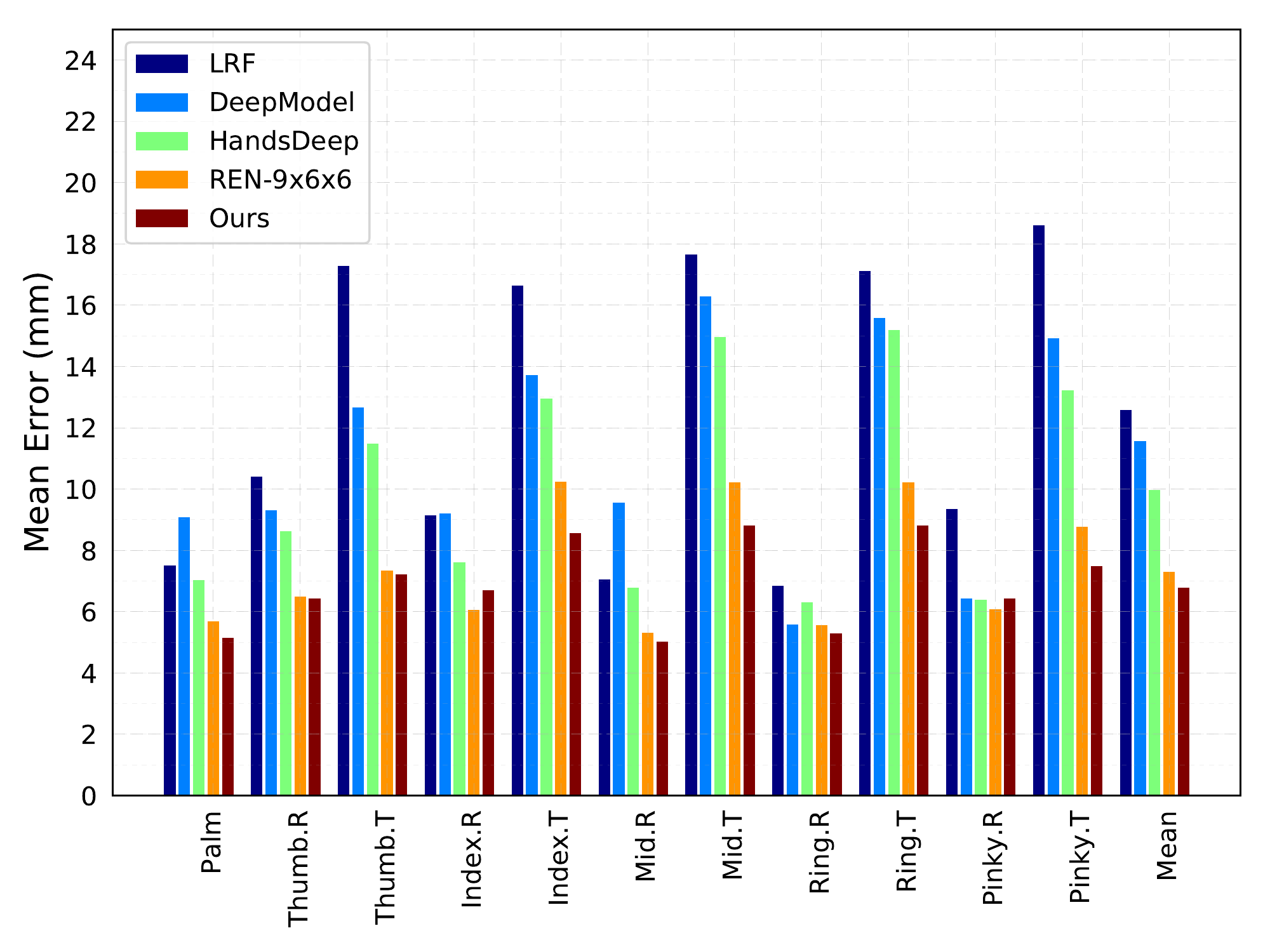}}
  \end{subfigure}%
\caption{Comparison of our approach with state-of-the-art methods on ICVL dataset.  Left: the proportion of good frames over different error thresholds. Right: per-joint errors.}
\label{fig:icvl-results}
\end{figure*}

\begin{figure*}[!tb]
  \centering
  \begin{subfigure}[t]{0.5\textwidth}
    \centering
    \centerline{\includegraphics[width=\linewidth]{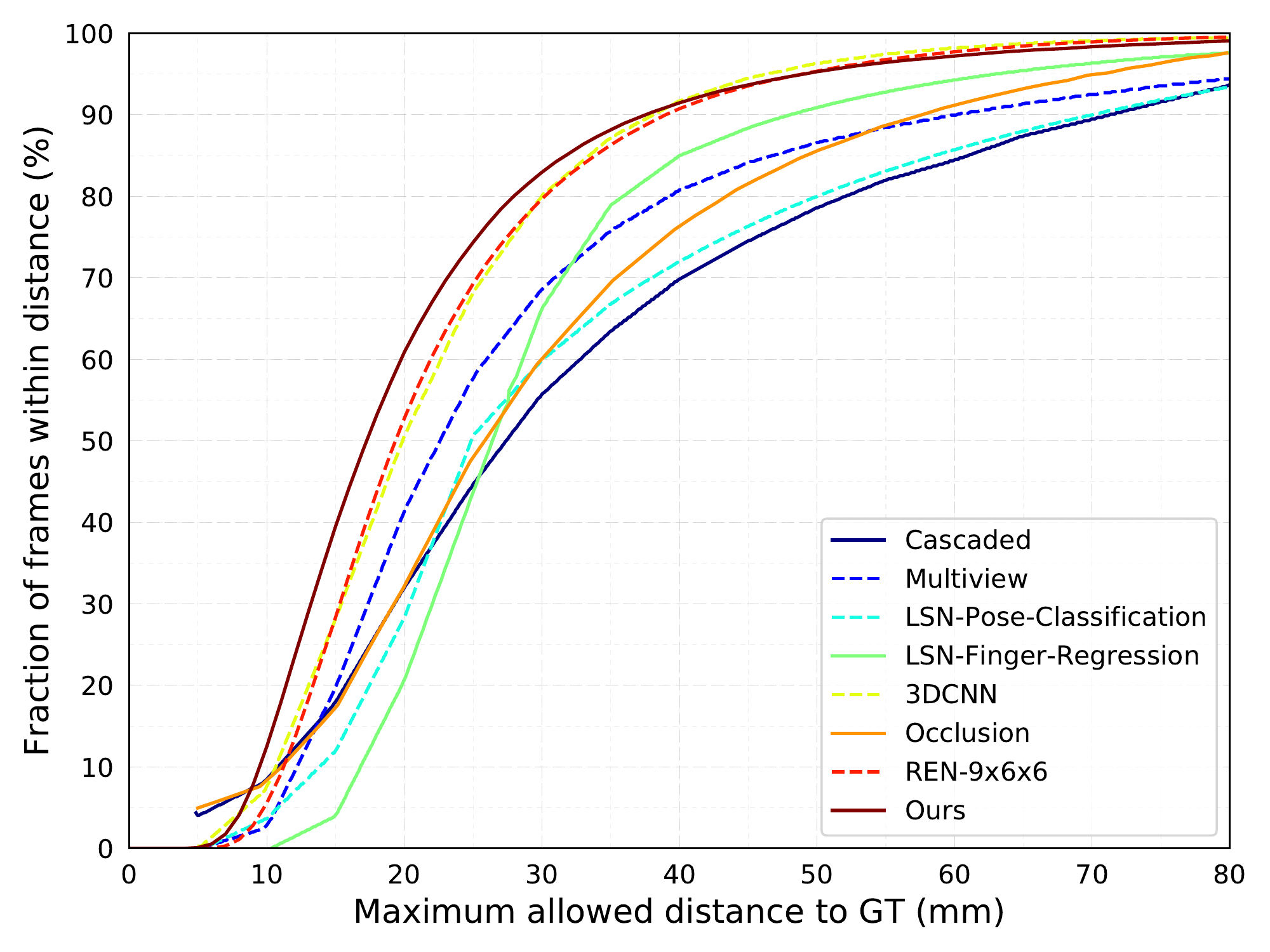}}
  \end{subfigure}%
  \begin{subfigure}[t]{0.5\textwidth}
    \centering
    \centerline{\includegraphics[width=\linewidth]{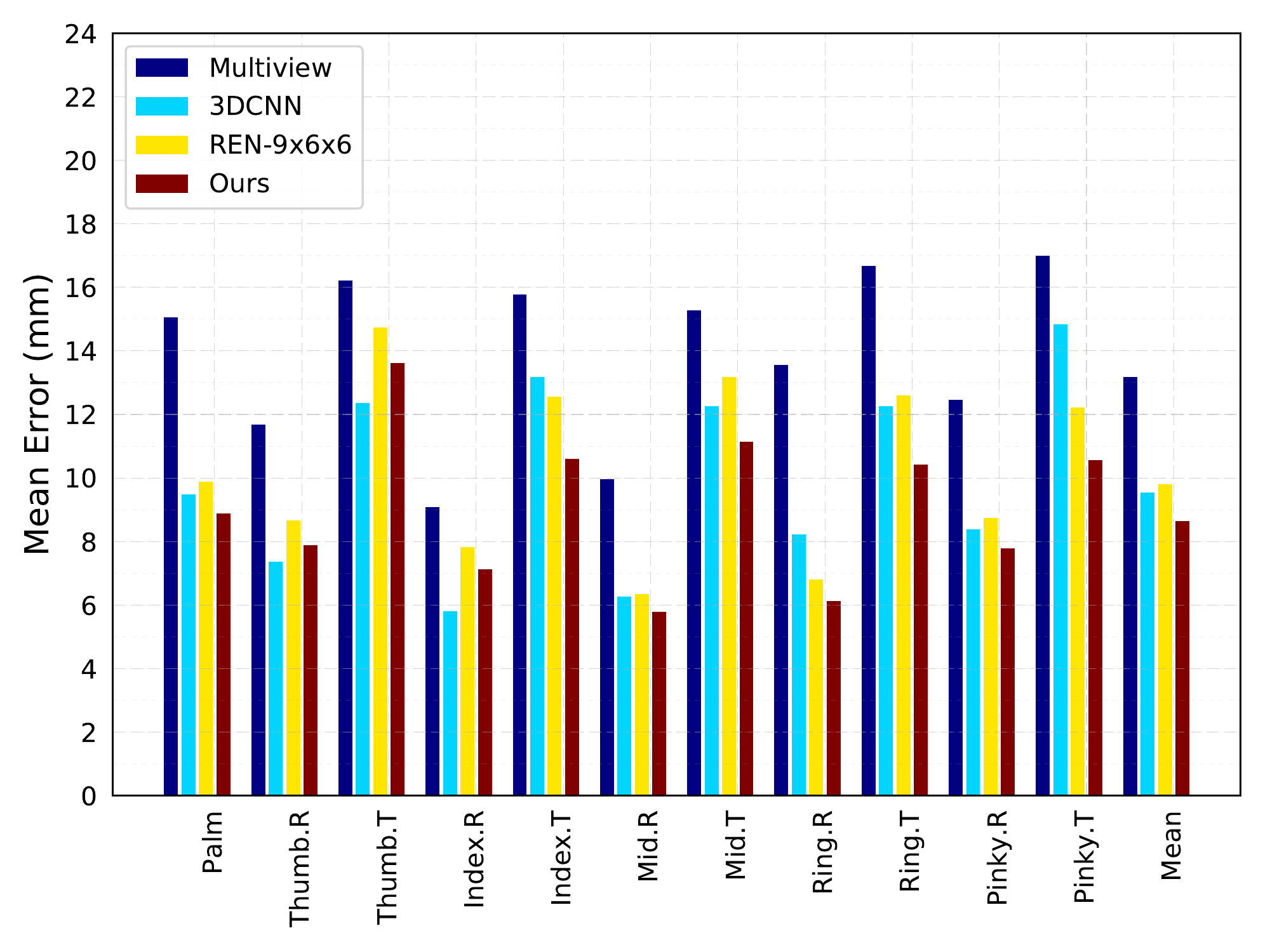}}
  \end{subfigure}%
\caption{Comparison of our approach with state-of-the-art methods on MSRA dataset. Left: the proportion of good frames
over different error thresholds. Right: per-joint errors.}
\label{fig:msra-results}
\end{figure*}

\begin{figure*}[!tb]
  \centering
  \begin{subfigure}[t]{0.5\textwidth}
    \centering
    \centerline{\includegraphics[width=\linewidth]{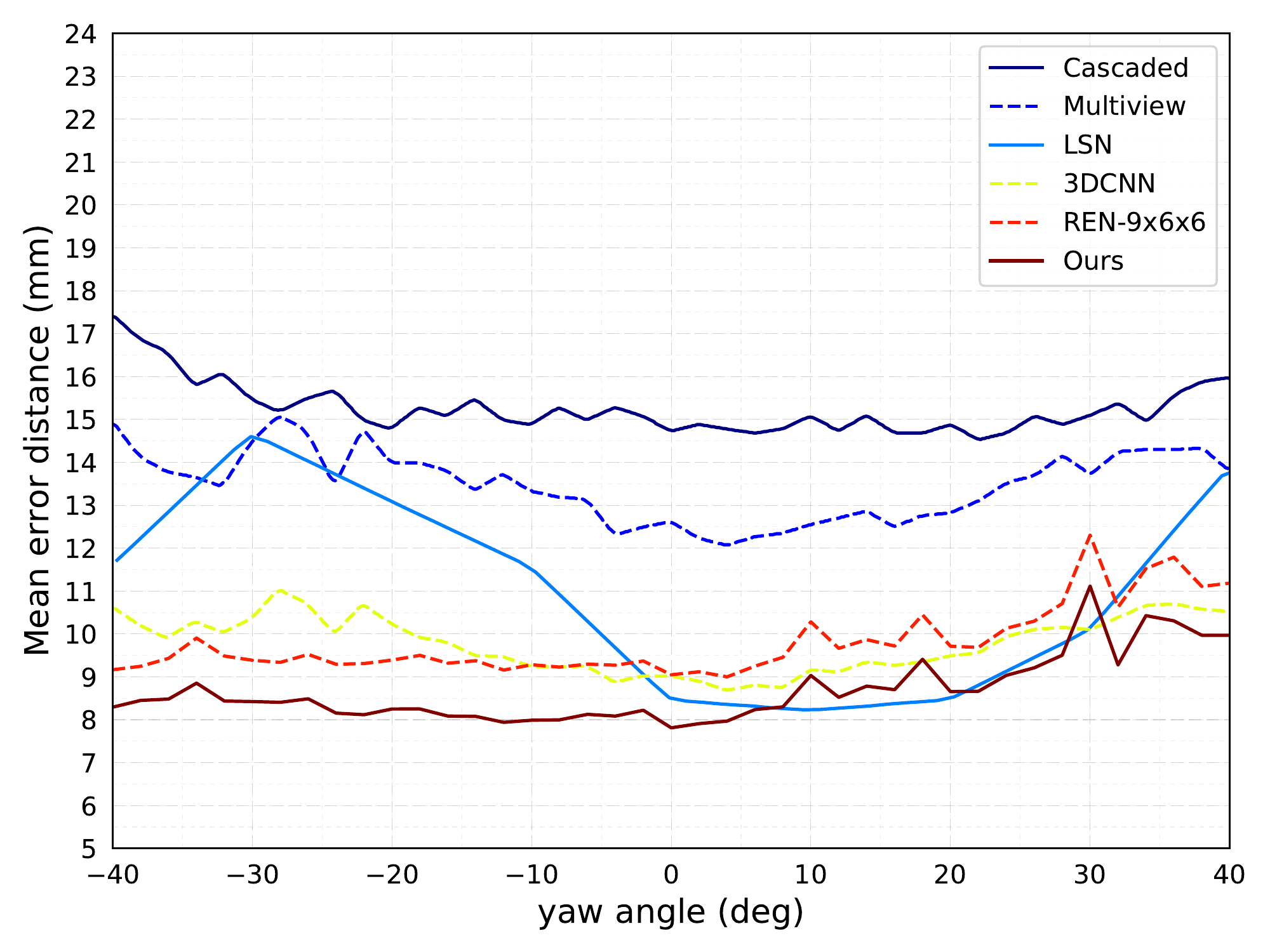}}
  \end{subfigure}%
  \begin{subfigure}[t]{0.5\textwidth}
    \centering
    \centerline{\includegraphics[width=\linewidth]{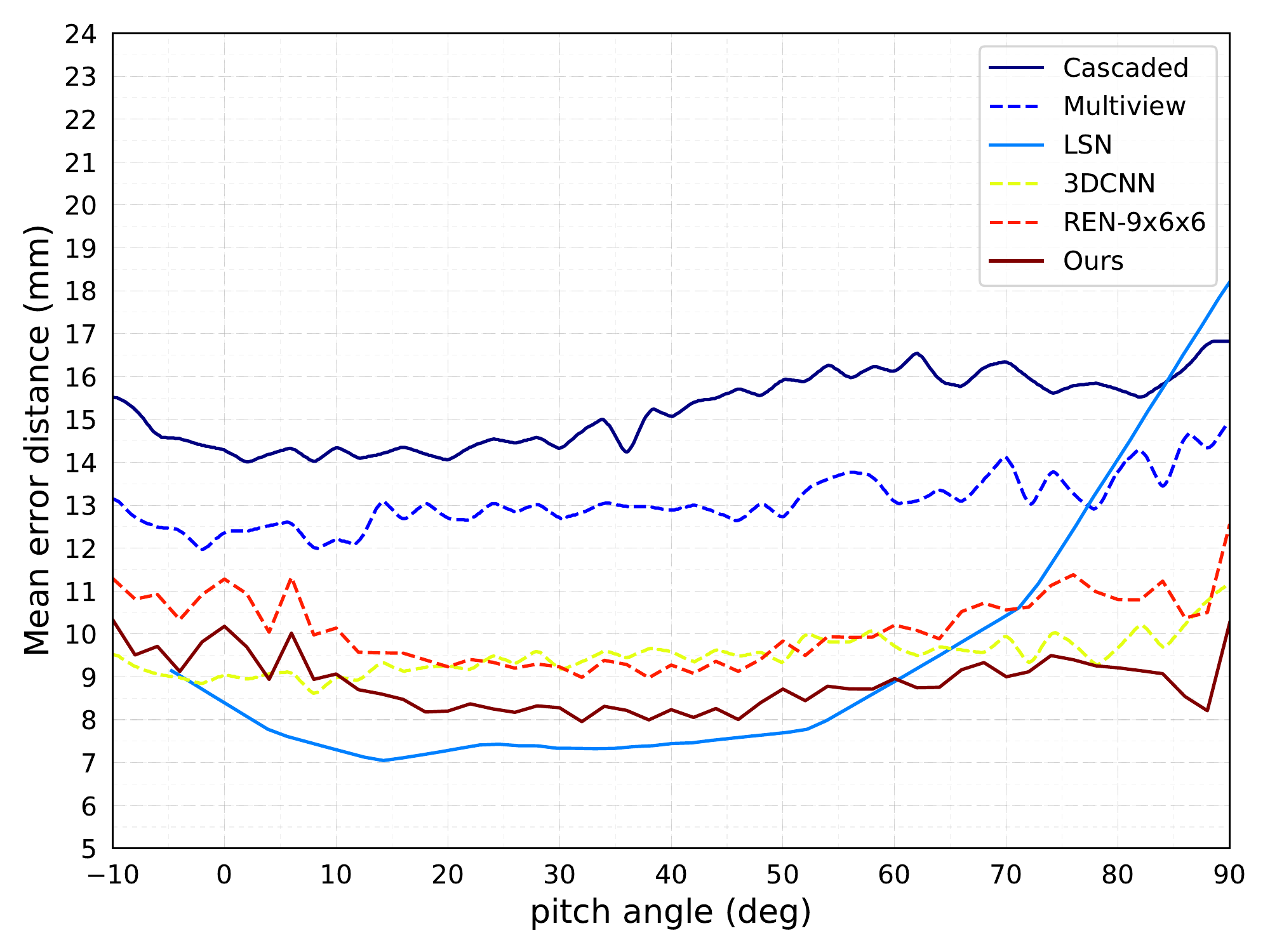}}
  \end{subfigure}%
\caption{Comparison of mean error distance over different yaw (left) and pitch (right) viewpoint angles on MSRA dataset.}
\label{fig:msra-results-angle}
\end{figure*}

\section{Experiments}
\label{sec:exp}
In this section, we will first introduce the datasets and evaluation metrics in the experiments. Afterwards we will evaluation our proposed method on three challenging public datasets: ICVL Hand Posture Dataset~\cite{tang2014latent}, NYU Hand Pose Dataset~\cite{tompson2014real} and MSRA Hand Pose Dataset~\cite{sun2015cascaded}. Finally we conduct extensive experiments for ablation study to discuss the effectiveness and robustness of different components of our proposed method.

\subsection{Datasets}
\paragraph{\textbf{ICVL Hand Posture Dataset}~\cite{tang2014latent}} This dataset was collected from $10$ different subjects using Intel's Creative Interactive Gesture Camera~\cite{melax2013dynamics}. In-plane rotations are applied to the collected samples and the final dataset contains $330k$ samples for training. There are totally $1596$ samples in the testset, including $702$ samples for test sequence A and $894$ samples for test sequence B. The annotation of hand pose contains $16$ joints, including $3$ joints for each finger and $1$ joint for the palm.
\paragraph{\textbf{NYU Hand Pose Dataset~\cite{tompson2014real}}} The NYU hand pose dataset was collected using three Kinects from different views. The training set contains $72757$ frames from $1$ subject and the testing set contains $8252$ frames from $2$ subjects, while one of the subjects in testing set doesn't appear in training set. The annotation of hand pose contains $36$ joints. Following the protocol of previous works~\cite{tompson2014real, oberweger2015hands, oberweger2015training, zhou2016model, guo2017region}, we only use frames from the frontal view and $14$ out of $36$ joints in evaluation.
\paragraph{\textbf{MSRA Hand Pose Dataset~\cite{sun2015cascaded}}} The MSRA hand pose dataset contains $76500$ frames from 9 different subjects captured by Intel's Creative Interactive Camera. The leave one subject out cross validation strategy is utilized for evaluation. The annotation of hand pose consists of $21$ joints, with $4$ joints for each finger and $1$ joint for the palm. This dataset has large viewpoint variation, which makes it a rather challenging dataset.

\subsection{Evaluation Metric}
There are two evaluation metrics widely used in hand pose estimation: per-joint errors and success rate. Denote $\{p_{ij}\}$ as the predicted joint locations of test frames, where $i$ is the index of frame and $j$ is the index of joint. $\{{p}^{gt}_{ij}\}$ is the corresponding groundtruth label. $N$ is the number of test frames and $J$ is the number of joints in a frame.
\paragraph{\textbf{Per-joint Errors}} Average euclidean distance between predicted joint location and groundtruth for each joint over all test frames. The error for the $j^{th}$ joint is calculated by:
\begin{equation}\label{eq-mean-error}
err_j = {\frac {\sum_{i}(\|p_{ij}-p^{gt}_{ij}\|)} {N}}
\end{equation}
Average joint error $err= {\frac {\sum_{j}err_j} {J}}$ is also used to evaluate the overall performance of hand pose estimation.

\paragraph{\textbf{Success Rate}} The fraction of good frames. A frame is considered as good if the maximum joint error of this frame is within a distance threshold $\tau$. The success rate for distance threshold $\tau$ is calculated as Eq.~\ref{eq:success-rate}.
\begin{equation}\label{eq:success-rate}
rate_{\tau} = \frac {\sum_{i}\mathbb{1}(\max_j(\|p_{ij}-p^{gt}_{ij}\|) \leq \tau)} {N}
\end{equation}
where $\mathbb{1}(cond)$ is an indicate function that equals to one if $cond$ is true and equals to zero otherwise.

\subsection{Comparison with State-of-the-Arts}

\begin{table}[!tb]
\centering
\caption{Quantitative evaluation of different methods on the benchmark NYU dataset for hand pose estimation task. We report 2D average pixel errors and 3D average joint errors in mm.}
\vspace{0.1cm}
\label{tab:table-nyu}
\begin{tabular}{lcc}
\hline
Methods& 3D error (mm) & 2D error (pixels) \\\hline
HandsDeep~\cite{oberweger2015hands} & 19.73 & 9.81 \\
Feedback~\cite{oberweger2015training} & 15.97 & 8.20\\
DeepModel~\cite{zhou2016model} & 16.90 & 8.76\\
Mask R-CNN~\cite{he2017mask} & 27.61 & 8.25\\
JTSC~\cite{fourure2017multi} & 16.80 & 8.02\\
Madadi \textit{et al.}~\cite{madadi2017end} & 15.60&- \\
Lie-X~\cite{Xu2017} & 14.51 & 7.48\\
REN (4x6x6)~\cite{guo2017region} & 13.39 & 6.78\\
REN (9x6x6)~\cite{wang2018region} & 12.69 & 6.32\\\hline
Ours & $\bm{11.81}$ & $\bm{5.53}$\\\hline
\end{tabular}
\end{table}

To demonstrate the effectiveness of our proposed method, we compare it against several state-of-the-art methods, including latent random forest (LRF)~\cite{tang2014latent}, DeepPrior with refinements (HandsDeep)~\cite{oberweger2015hands}, cascaded hand pose regression (Cascaded)~\cite{sun2015cascaded}, feedback loop (Feedback)~\cite{oberweger2015training}, deep hand model (DeepModel)~\cite{zhou2016model}, Lie group based method (Lie-X)~\cite{Xu2017}, multi-view CNN (Multiview)~\cite{ge2016robust}, 3D-CNN based method (3DCNN)~\cite{ge2017threedcnn} , CrossingNets~\cite{wan2017crossing}, local surface normals (LSN)~\cite{wan2016hand}, occlusion aware method (Occlusion)~\cite{madadi2017occlusion}, JTSC~\cite{fourure2017multi}, global to local CNN (Madadi \textit{et al.})~\cite{madadi2017end} and region ensemble network with $9\times6\times6$ region setting (REN-9x6x6)~\cite{wang2018region}.

It should be noted that some reported results of state-of-the-art methods are calculated using the predicted labels that are available online~\cite{tang2014latent, oberweger2015training, zhou2016model, guo2017region, Xu2017, wang2018region} and others are estimated from the figures and tables of the original papers~\cite{sun2015cascaded, wan2016hand, ge2016robust, ge2017threedcnn, wan2017crossing, madadi2017occlusion, madadi2017end}.

We also compare our method with Mask R-CNN~\cite{he2017mask} due to its impressive performance on RGB human pose estimation. For fair comparison, we first crop the depth images and resize them into $96\times96$, which is the same preprocessing as our proposed method. We use similar setting with human pose estimation task in~\cite{he2017mask} that exploits ResNet-50-FPN as the backbone network. To adopt Mask R-CNN for depth-based hand pose estimation, we first use Mask R-CNN to detect 2D hand pose in image coordinates and then infer depth values from the original depth images to recover 3D hand pose. To alleviate the impact of noises and holes in depth images, the inferred depth values are constrained within the 3D cube of hand and valid depth values from 9-neighbours are averaged to get the final depth coordinate.

On NYU dataset, we compare our proposed method with~\cite{oberweger2015hands, oberweger2015training, zhou2016model, sinha2016deephand, Xu2017, wan2017crossing, ge2016robust, ge2017threedcnn, wang2018region, he2017mask}. The success rate with respect to the worse case criteria and per-joint errors are given in Figure~\ref{fig:nyu-results}. As shown in the figure, our proposed outperforms all state-of-the-art methods. We further compare the overall 2D and 3D mean joint error in Table~\ref{tab:table-nyu}. Our method obtain $0.88mm$ 3D error decrease compared with existing best performance by REN~\cite{wang2018region}.
Mask R-CNN performs 2D keypoint detection and the post-processing is used to lift 2D pose to 3D pose. It achieves comparable 2D error with prior methods. Nevertheless, our Pose-REN outperforms Mask R-CNN and reduces the 2D error by $2.7$ pixels.

On ICVL dataset, we compare our proposed method against~\cite{tang2014latent, zhou2016model, oberweger2015hands, wan2017crossing, wan2016hand, wang2018region}. Results in Figure~\ref{fig:icvl-results} demonstrate that our proposed method outperforms all other methods with a large margin. Compared with REN~\cite{wang2018region}, our method reduces the mean error by $0.514mm$, which is a $7.04\%$ relative improvement.

On MSRA dataset, we compare with several state-of-the-art methods~\cite{sun2015cascaded, ge2016robust, ge2017threedcnn, wan2016hand, madadi2017occlusion, wang2018region}. The success rate with respect to maximum allowed threshold and per-joint errors are shown in Figure~\ref{fig:msra-results}. Our method achieves the best performance among all evaluated methods. Following the protocol of previous works~\cite{sun2015cascaded}, we also report the mean joint errors distributed over yaw and pitch viewpoint angles, as shown in Figure~\ref{fig:msra-results-angle}. Our method achieves the smallest errors in almost all angles. It should be noted that the LSN~\cite{wan2016hand} get slightly smaller errors when the yaw or pitch angle is relatively small. However, the performance of LSN decreases rapidly when the viewpoint becomes larger. These results demonstrate that our method is much more robust to viewpoint changes, which is a quite challenging problem in hand pose estimation.

The fraction of good frames of our method decreases slightly compared with REN~\cite{wang2018region} when the errors are larger than around $30mm$. This is mainly due to worse initial pose for these challenging samples.
When regarding to the per-joint errors, our method achieves the best performance among all compared methods.

\subsection{Ablation Study}
\label{sec:ablation}
In this section we will provide extensive experiments to discuss the contributions of different components of our method and the effect of some parameters.

\begin{figure*}[!tb]
  \centering
  \begin{subfigure}[t]{0.5\textwidth}
    \centering
    \centerline{\includegraphics[width=\linewidth]{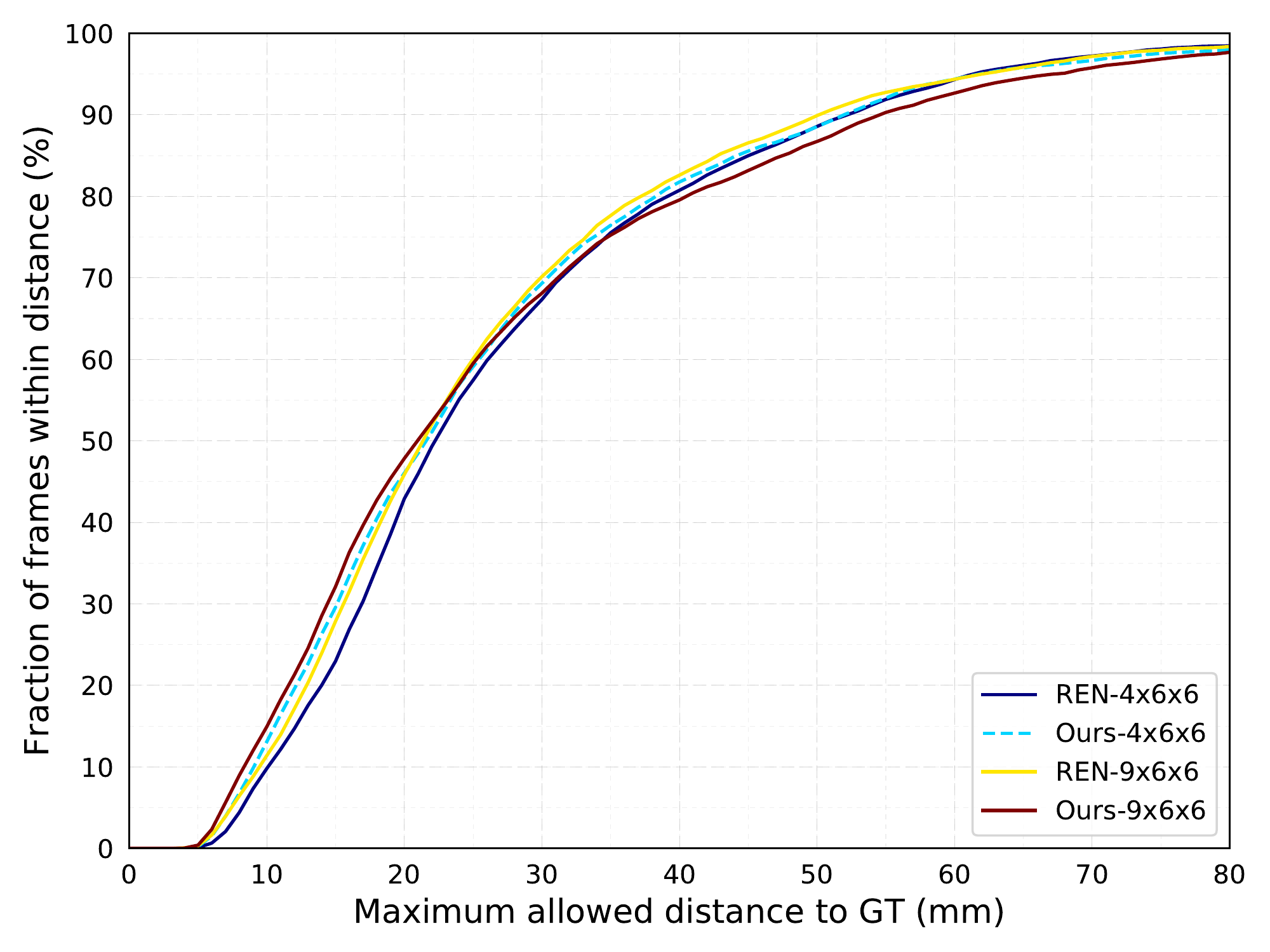}}
  \end{subfigure}%
  \begin{subfigure}[t]{0.5\textwidth}
    \centering
    \centerline{\includegraphics[width=\linewidth]{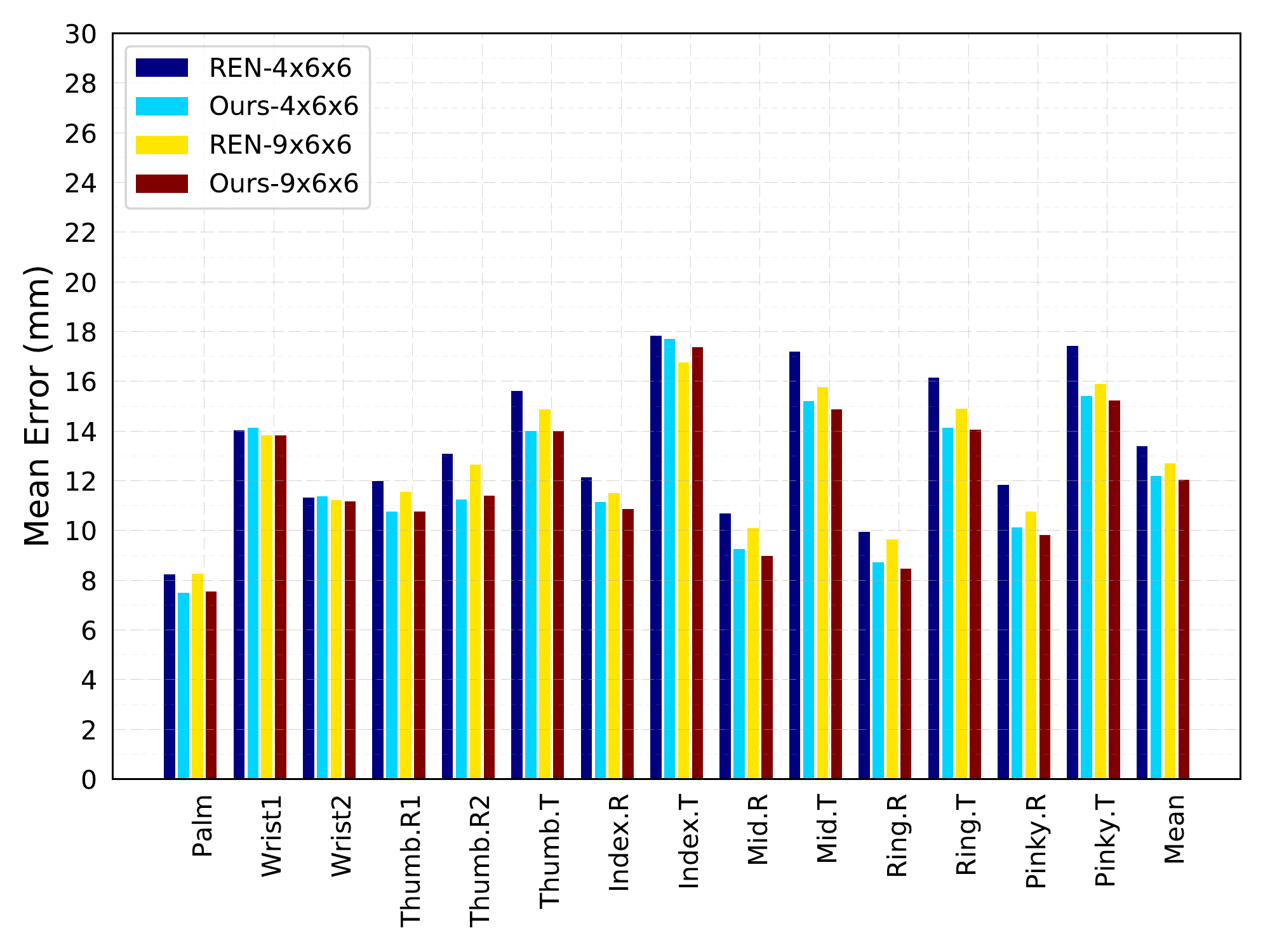}}
  \end{subfigure}%
\caption{Effect of pose guided region ensemble by comparing our method against grid region ensemble (REN~\cite{guo2017region}).  Left: the proportion of good frames over different error thresholds. Right: per-joint errors.}
\label{fig:nyu-ablation-guided}
\end{figure*}

\subsubsection{Effect of the Number of Iteration $T$}
\label{sec:ablation-iter}
\begin{figure}[!tb]
\centering
\centerline{\includegraphics[width=\linewidth]{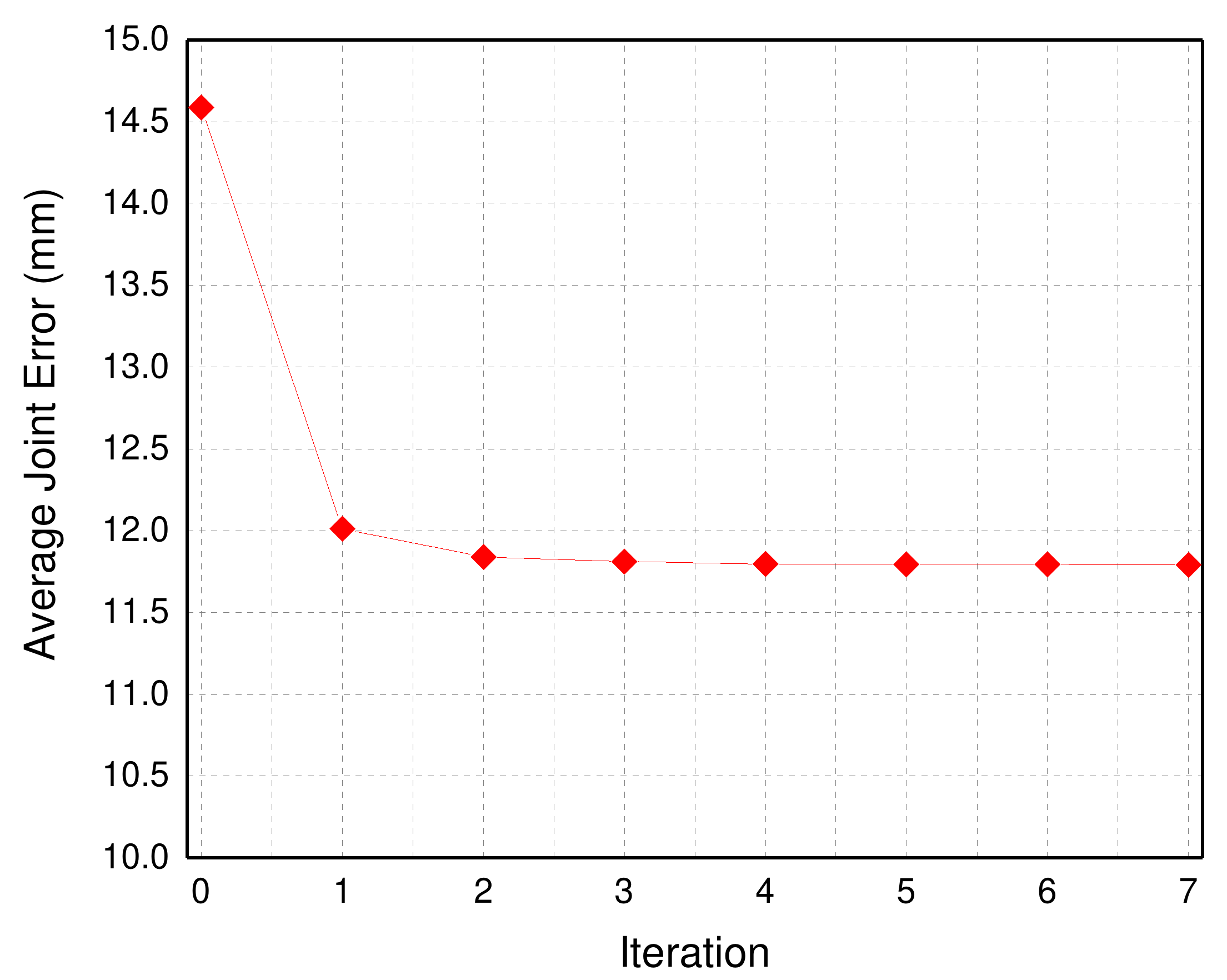}}
\caption{Effect of the number of iteration on NYU dataset.}
\label{fig:nyu-ablation-iteration}
\end{figure}

First we will discuss how the number of iteration $T$ affects the performance. The average joint errors on NYU dataset with using the different number of iterations are shown in Figure~\ref{fig:nyu-ablation-iteration}. The error for iteration $0$ is the result of the initialization. After one iteration, the error drops rapidly. As the iteration increases, the error becomes stable and finally converges. To better balance the computation complexity and performance, we choose the number of iteration as $T=3$.

\subsubsection{Effect of Pose Guided Region Extraction}
\label{sec:ablation-pose-guided}
One of the contributions of our proposed method is to extract feature regions under the guidance of hand pose from previous stage. We will show whether this strategy helps to improve the performance of hand pose estimation. In REN ~\cite{guo2017region}, feature regions are extracted using a uniformly distributed grid. We report the performances of our method that only adopts one iteration and sets the number of regions and the size of regions the same as REN-4x6x6~\cite{guo2017region} and REN-9x6x6~\cite{wang2018region}. Under such experimental settings, the number of parameters of our method and REN are the same, which ensures fair comparison. The first number in the suffix indicates the number of regions and the last two numbers represent the size of regions. Specifically, we use the palm joint, the root joint of thumb, middle, pinky finger in 4x6x6 setting (denoted as Our-4x6x6) and use all joints except for two joints in thumb finger and the tip joint of pinky finger in 9x6x6 setting (denoted as Our-9x6x6). The success rate curve and per-joint errors on NYU dataset are shown in Figure~\ref{fig:nyu-ablation-guided}. With different region settings, our method both performs better than REN that adopts grid region ensemble, indicating the contributions of pose guided region extraction strategy.

\begin{figure*}[!tb]
  \centering
  \begin{subfigure}[t]{0.5\textwidth}
    \centering
    \centerline{\includegraphics[width=\linewidth]{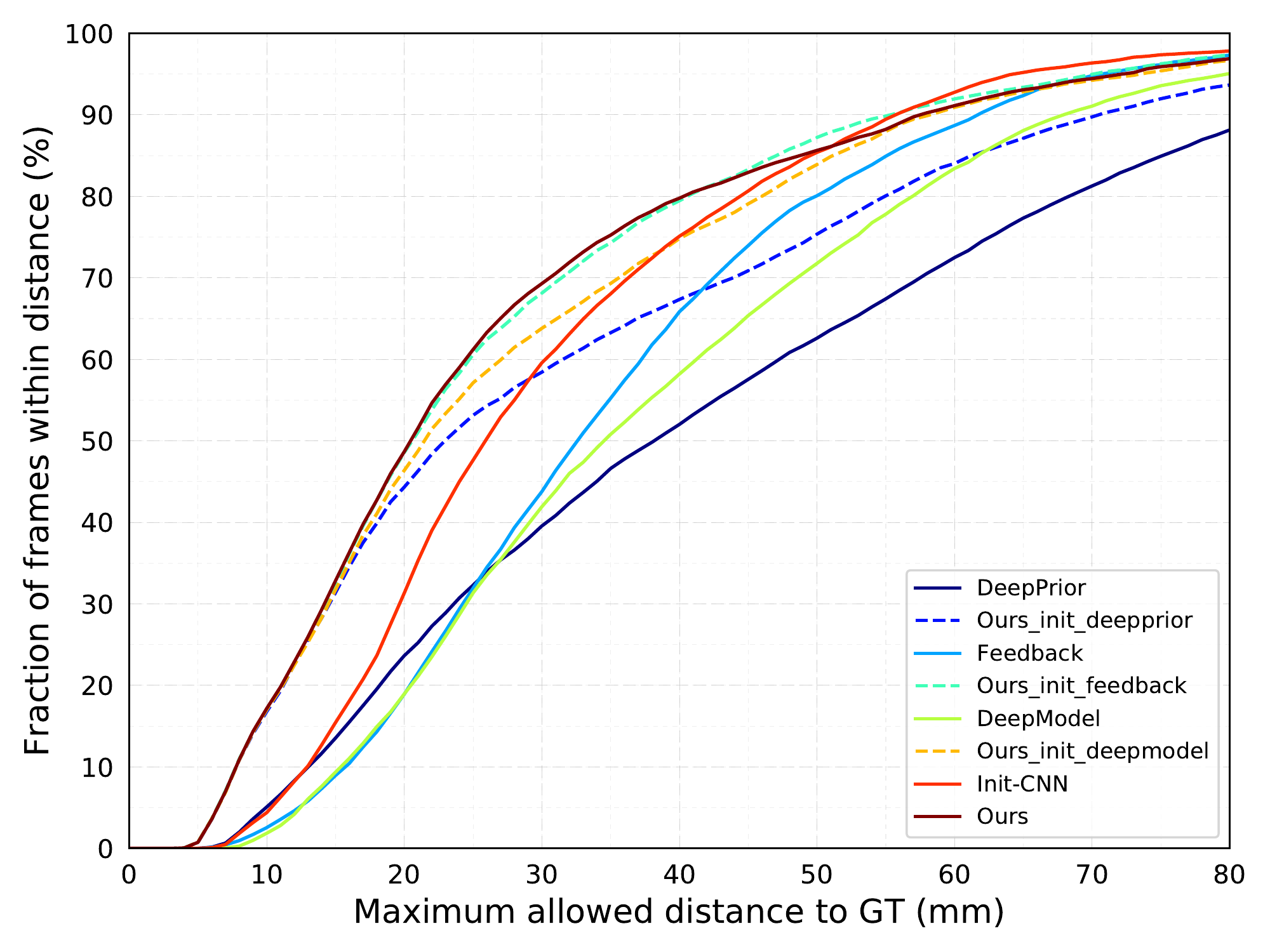}}
  \end{subfigure}%
  \begin{subfigure}[t]{0.5\textwidth}
    \centering
    \centerline{\includegraphics[width=\linewidth]{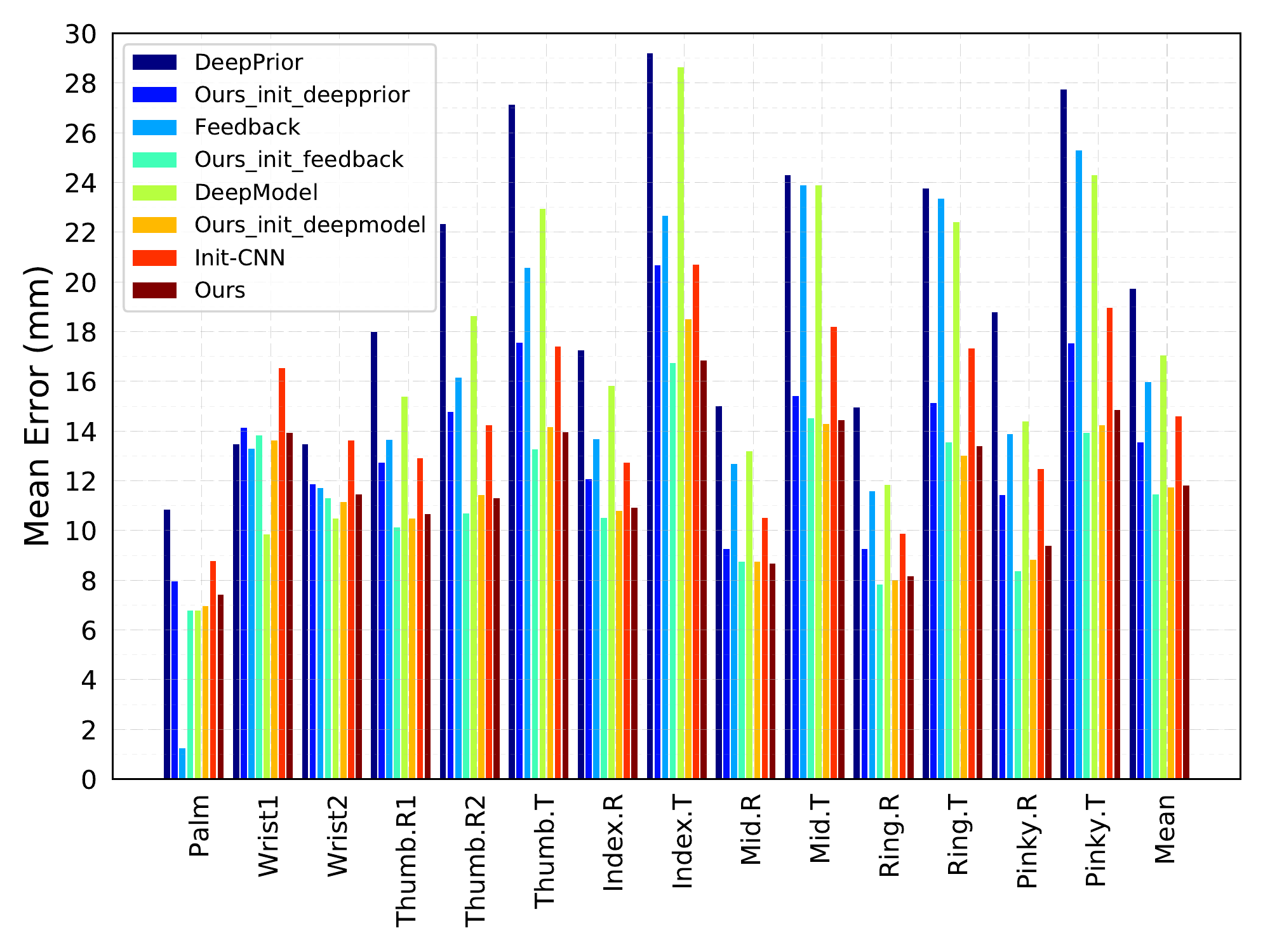}}
  \end{subfigure}%
\caption{Performance of our model with different initial hand pose used in inference phase on NYU dataset. Left: the proportion of good frames
over different error thresholds. Right: per-joint errors.}
\label{fig:nyu-ablation-initialization}
\end{figure*}

\begin{figure*}[!tb]
  \centering
  \begin{subfigure}[t]{0.5\textwidth}
    \centering
    \centerline{\includegraphics[width=\linewidth]{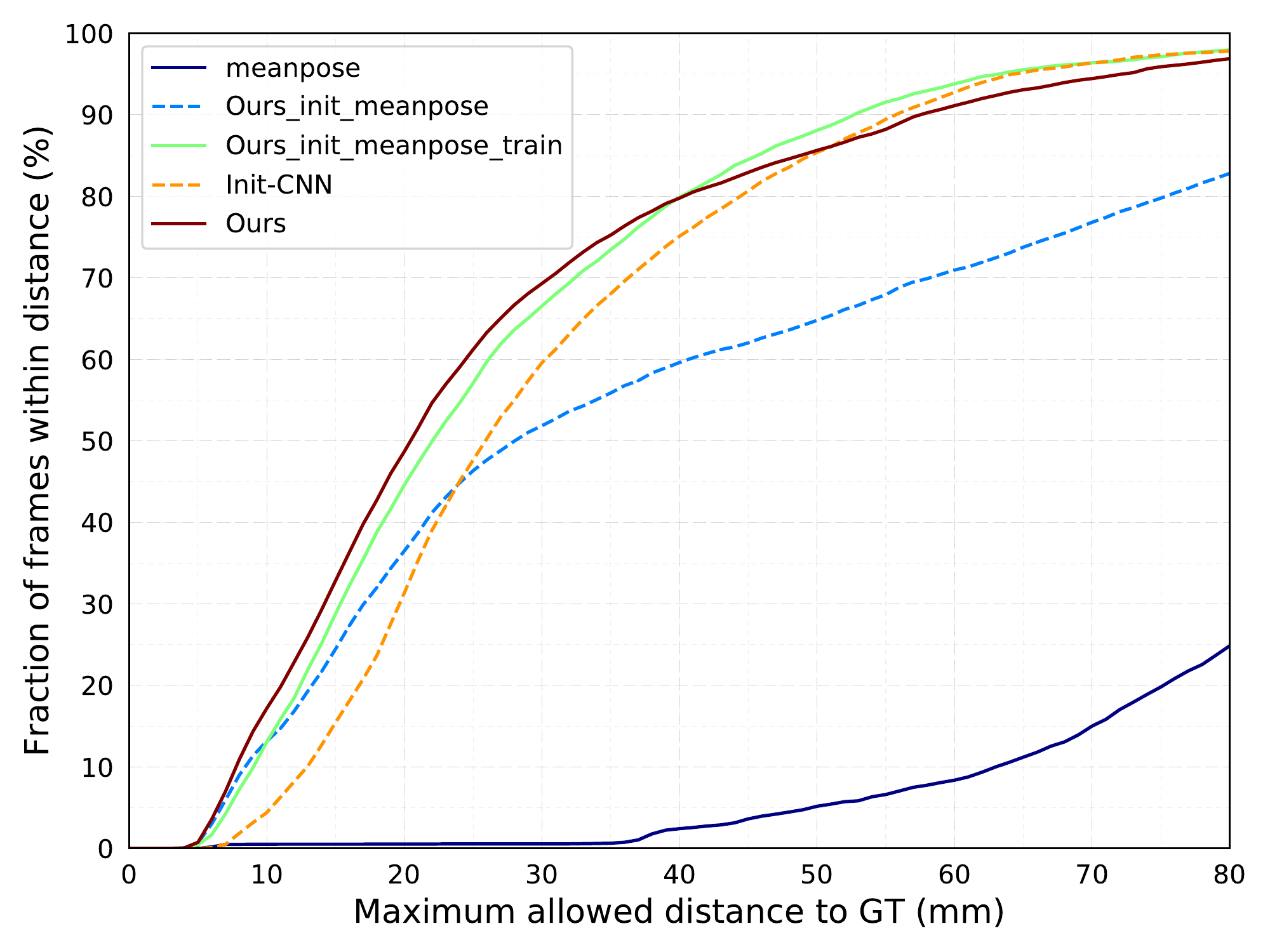}}
  \end{subfigure}%
  \begin{subfigure}[t]{0.5\textwidth}
    \centering
    \centerline{\includegraphics[width=\linewidth]{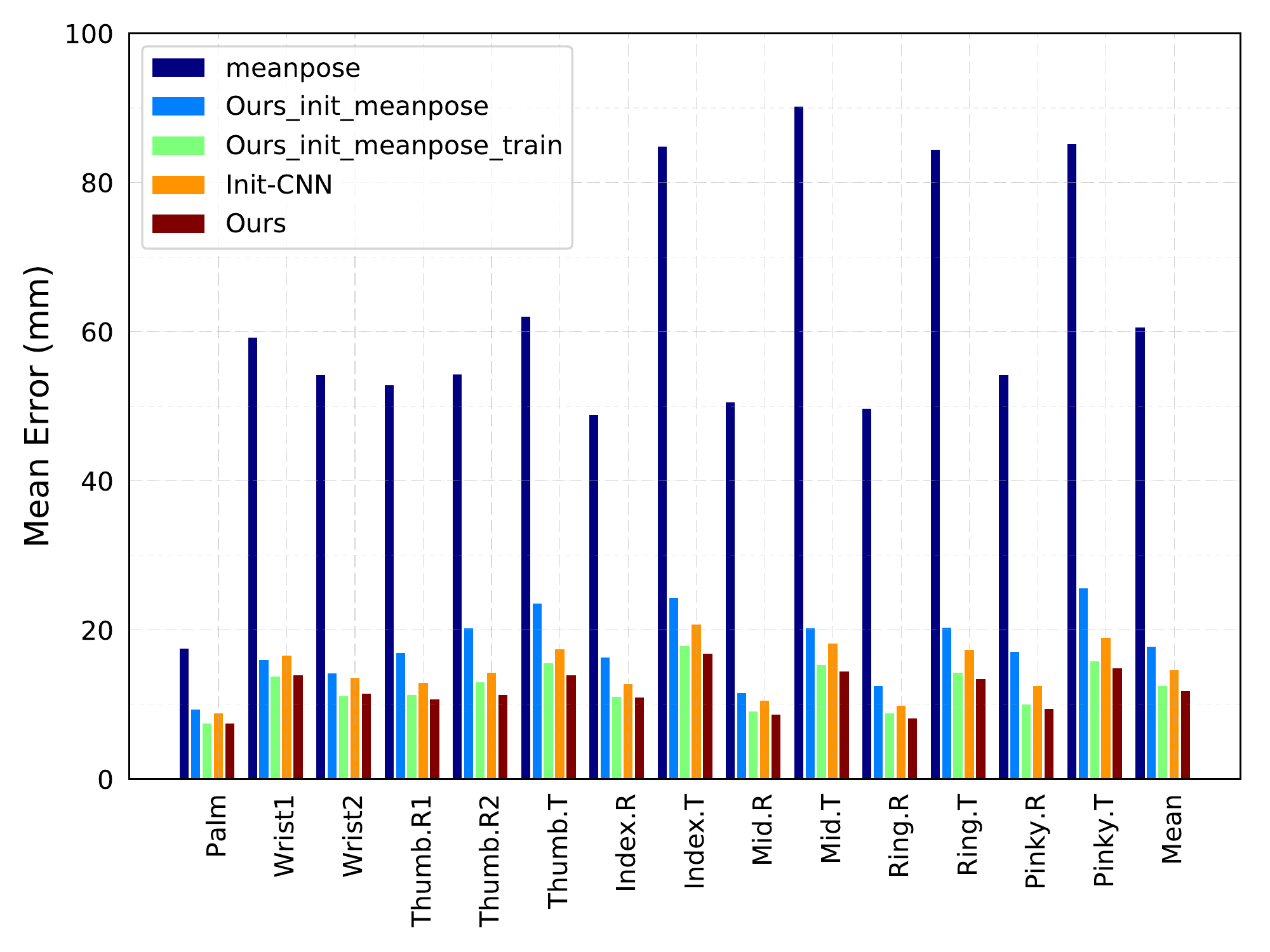}}
  \end{subfigure}%
\caption{Performance of our method when using mean pose as the initialization on NYU dataset. Left: the proportion of good frames
over different error thresholds. Right: per-joint errors.}
\label{fig:nyu-ablation-initialization-meanpose}
\end{figure*}

\subsubsection{Effect of Structured Region Ensemble}
We will demonstrate the effectiveness of another component of our proposed method: the hierarchically structured region ensemble. We compare our method with a network (denoted as $Ours\_w/o\_structure$) that use two simple $fc$ layers as is adopted in REN~\cite{guo2017region} instead of hierarchical $fc$ layers. For fair comparison, we set the dimensions of the two $fc$ layers as $2304$ and $2048$ respectively to ensure the similar number of parameters between our method and $Ours\_w/o\_structure$. The mean joint errors on NYU, ICVL and MSRA dataset are shown in Table~\ref{fig:nyu-ablation-structure}. It can be seen that our method performs better than $Ours\_w/o\_structure$, which illustrates the effectiveness of the hierarchically structured region ensemble strategy.

\begin{table}[!tb]
\centering
\caption{Comparing average joint errors of our method with and without structured region ensemble strategy on three datasets. The numbers in the brackets indicate the percentages of error reduction.}
\vspace{0.1cm}
\label{fig:nyu-ablation-structure}
\begin{tabular}{ccc}
\hline
Dataset & Ours w/o structure& Ours (mm)\\
 & (mm) & \\\hline
NYU~\cite{tompson2014real} & 11.869 & $\bm{11.811} (-0.5\%)$ \\
ICVL~\cite{tang2014latent} & 6.932 & $\bm{6.793} (-2.0\%)$ \\
MSRA~\cite{sun2015cascaded} & 8.728 & $\bm{8.649} (-0.9\%)$ \\\hline
\end{tabular}
\end{table}

\subsubsection{Effect of the Initialization}
\label{sec:ablation-init}
In this section we will demonstrate the robustness of our proposed method over different initializations.
Our proposed method builds upon the cascaded framework, which takes an initial hand pose as input and iteratively refine the results. To explore the impact of initialization for our methods, we conduct several experiments on NYU dataset with different initializations.

Firstly we will discuss the impact of initialization in inference phase. Specifically, we chose four methods as initialization: Init-CNN (which is proposed in~\cite{guo2017region} as a baseline network and also adopted as the initialization of our method),  DeepPrior~\cite{oberweger2015hands}, Feedback~\cite{oberweger2015training}, DeepModel~\cite{zhou2016model}.
The results of different initializations and refined results (denoted as, e.g. $Ours\_init\_deepprior$) are shown in Figure~\ref{fig:nyu-ablation-initialization}. We can observe that our method can considerably boost the performances of the initializations. Even with some rather rough initialization (e.g. DeepPrior), the refined results boosted by our Pose-REN are quite competitive. With other better initializations (Feedback, DeepModel), the final results are similar to our method, even if their initializations are slightly worse than ours. These results indicate the robustness over initializations of our method.
It should be noted that the model used above were trained using the samples with our initialization (Init-CNN). We used different initializations in inference to get the results above. Therefore, the results above also demonstrate the generalization of our model.

Furthermore, we consider the case that uses a natural pose (denoted as $meanpose$) as initialization and discuss which performance can be expected. We used the model that was trained on our initialization to refine the hand pose with $meanpose$ as the initialization. As shown in Figure~\ref{fig:nyu-ablation-initialization-meanpose}, the initial meanpose is very poor and the results are boosted by adopting our method. We empirically find that the performance converges after $10$ stages ($Ours\_init\_meanpose$), resulting the average joint error of $17.708mm$, which is comparable with some state-of-the-are methods, as shown in Table~\ref{tab:table-nyu}.
We further trained a model using the $meanpose$ as initialization and report the refined results ($Ours\_init\_meanpose\_train$) in Figure~\ref{fig:nyu-ablation-initialization-meanpose}. It can be seen that the results are quite close to those of our method that uses a better initialization, which indicates that our proposed method is robust to different initializations.

As discussed above, the model trained on our initialization greatly generalize to other initializations. Furthermore, for a very poor initialization, our proposed method can still obtain satisfying results by training a model using this initialization.

\subsection{Qualitative Results}
\label{sec:qualitative}
\begin{figure}[!tb]
\centering
\centerline{\includegraphics[width=0.9\linewidth]{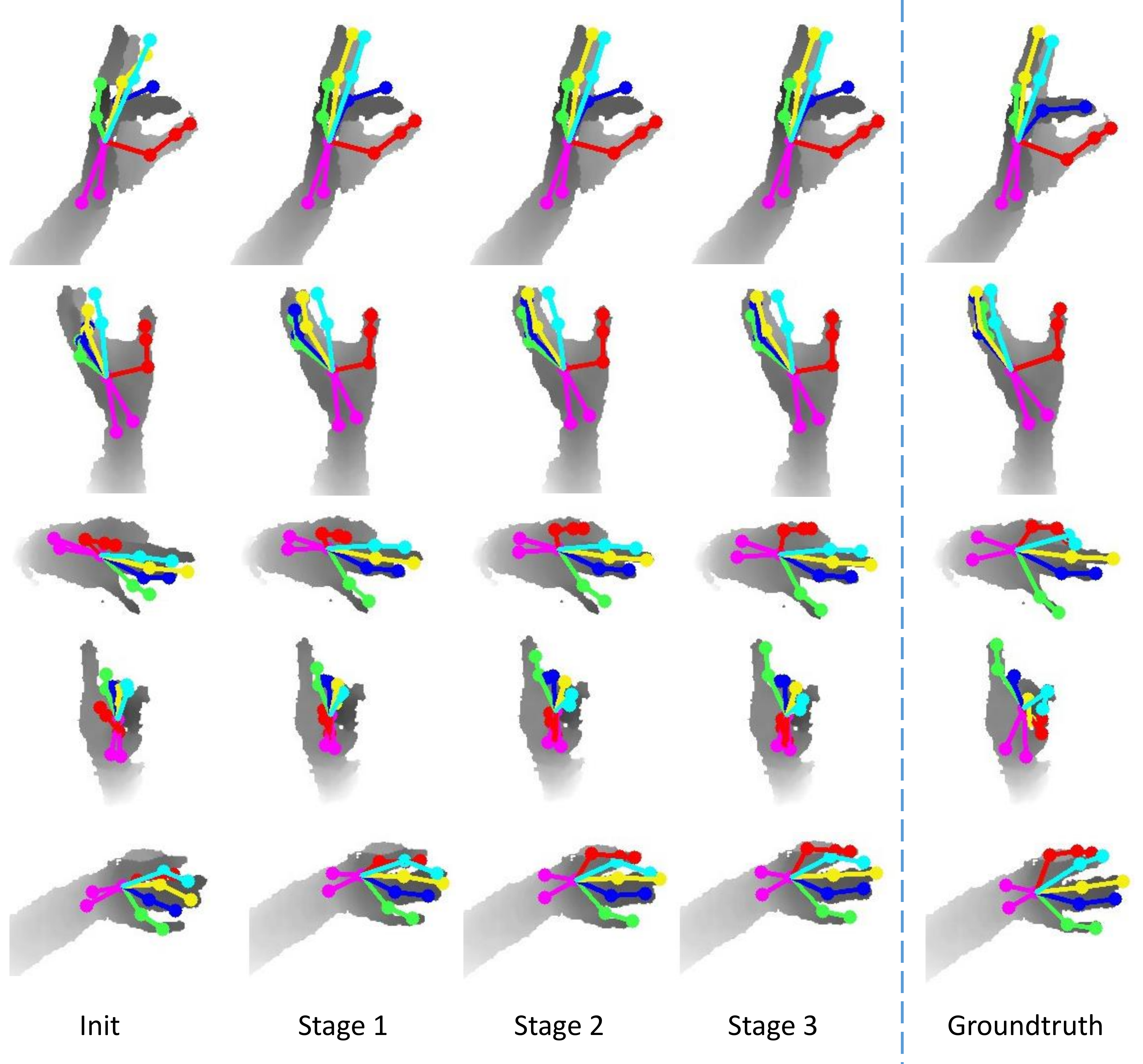}}
\caption{Qualitative results on NYU dataset of different stages.}
\label{fig:qualitative-cascaded}
\end{figure}

Figure~\ref{fig:qualitative-cascaded} shows some examples of the iterative process on NYU dataset. The first column shows the results of the initialize hand pose, the second to fourth columns show the refined results on stage $1 - 3$. The rightmost column is the groundtruth annotation. Our method gradually improves the estimated hand pose and obtains accurate results after several iterations.

\begin{figure*}[!htb]
\centering
\centerline{\includegraphics[width=\linewidth]{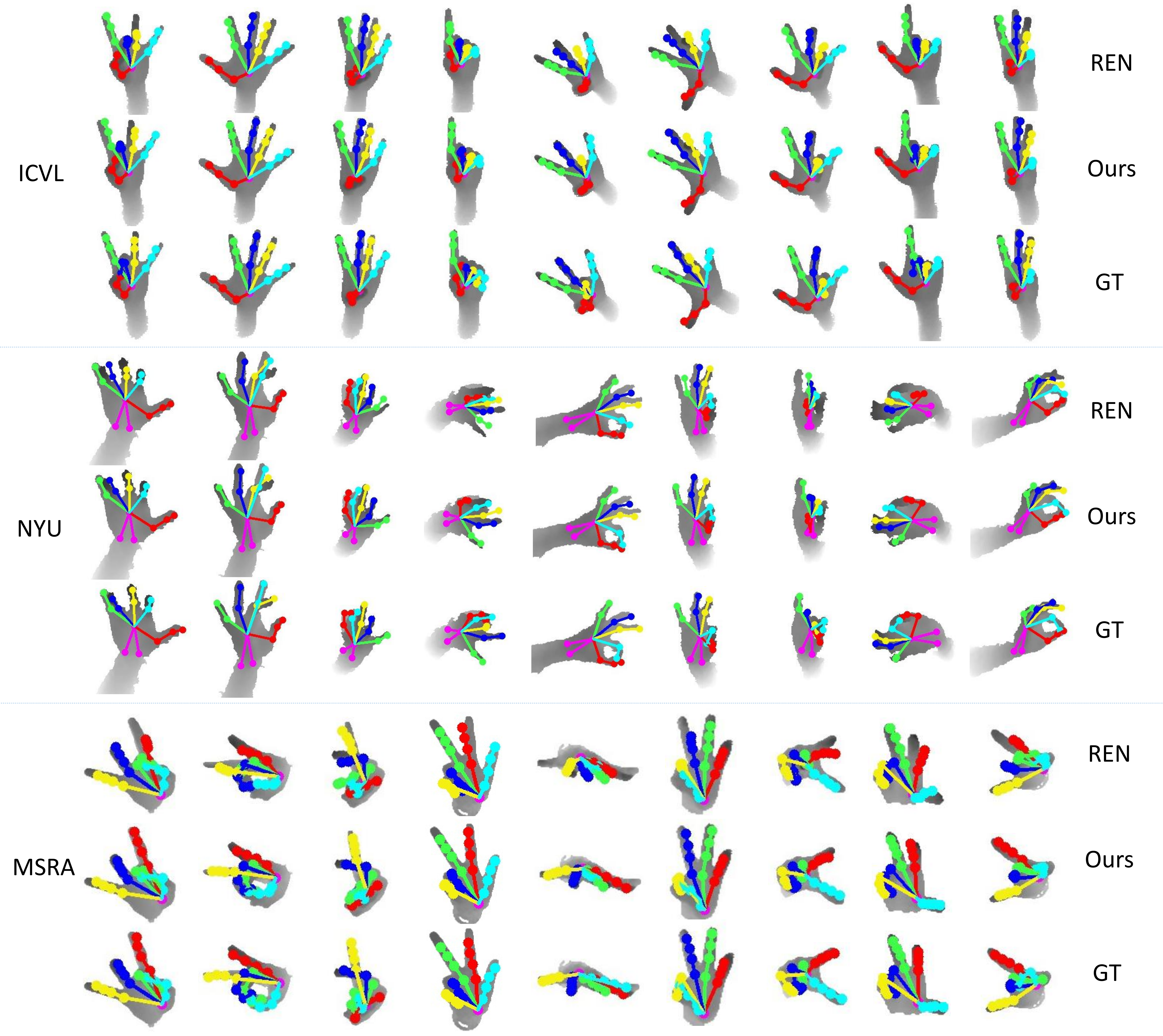}}
\caption{Qualitative results. For each dataset, three rows show the results from region ensemble network (REN-9x6x6)~\cite{wang2018region}, our method (Ours) and groundtruth (GT) respectively.}
\label{fig:qualitative-results}
\end{figure*}

Some qualitative results on three datasets can be seen in Figure~\ref{fig:qualitative-results}. For each dataset, the first row represents the results of REN-9x6x6~\cite{wang2018region}, the second row shows the results of our proposed method and the third row is the groundtruth. It can be seen that our method performs better than REN even in some challenging samples.

\section{Conclusion}
\label{sec:conclusion}
In this paper we propose a novel method called pose guided structured region ensemble network (Pose-REN) for accurate 3D hand pose estimation from a single depth image. Our method extracts regions from the feature maps under the guidance of an initially estimated hand pose to attain more optimal and representative features. Feature regions are then integrated hierarchically by adopting a tree-like structured connection that models the topology of hand joints. Our method iteratively refines the hand pose to obtain the final estimated results. Experiments on public hand pose datasets demonstrate that our proposed method outperforms all state-of-the-art methods.
In our future work, we intend to further improve our method for robust and accurate 3D hand pose estimation when hands are interacting with other hands or objects. We would like to research on integrating hand detection and hand pose estimation into a unified framework, based on Faster R-CNN\cite{ren2015faster} or Mask R-CNN~\cite{he2017mask} etc. It will also be interesting to apply our proposed method for more articulated pose estimation tasks, like human pose estimation and face alignment.

\section*{Acknowledgments}
This work was partially supported by State High-Tech Research and Development Program of China (863 Program) [grant number 2015AA016304].

\bibliography{mybibfile}

\begin{thebibliography}{10}
\expandafter\ifx\csname url\endcsname\relax
  \def\url#1{\texttt{#1}}\fi
\expandafter\ifx\csname urlprefix\endcsname\relax\def\urlprefix{URL }\fi
\expandafter\ifx\csname href\endcsname\relax
  \def\href#1#2{#2} \def\path#1{#1}\fi

\bibitem{erol2007vision}
A.~Erol, G.~Bebis, M.~Nicolescu, R.~D. Boyle, X.~Twombly, Vision-based hand
  pose estimation: A review, Computer Vision and Image Understanding 108~(1)
  (2007) 52--73.

\bibitem{chen2017motion}
X.~Chen, H.~Guo, G.~Wang, L.~Zhang, Motion feature augmented recurrent neural
  network for skeleton-based dynamic hand gesture recognition, in: Image
  Processing (ICIP), 2017 24th IEEE International Conference on, IEEE, 2017,
  pp. 2881--2885.

\bibitem{de2016skeleton}
Q.~De~Smedt, H.~Wannous, J.-P. Vandeborre, Skeleton-based dynamic hand gesture
  recognition, in: Proceedings of the IEEE Conference on Computer Vision and
  Pattern Recognition Workshops, 2016, pp. 1--9.

\bibitem{supancic2015depth}
J.~S. Supancic, G.~Rogez, Y.~Yang, J.~Shotton, D.~Ramanan, Depth-based hand
  pose estimation: data, methods, and challenges, in: Proceedings of the IEEE
  International Conference on Computer Vision, 2015, pp. 1868--1876.

\bibitem{tompson2014real}
J.~Tompson, M.~Stein, Y.~Lecun, K.~Perlin, Real-time continuous pose recovery
  of human hands using convolutional networks, ACM Transactions on Graphics
  (TOG) 33~(5) (2014) 169.

\bibitem{tang2015opening}
D.~Tang, J.~Taylor, P.~Kohli, C.~Keskin, T.-K. Kim, J.~Shotton, Opening the
  black box: Hierarchical sampling optimization for estimating human hand pose,
  in: Proceedings of the IEEE International Conference on Computer Vision,
  2015, pp. 3325--3333.

\bibitem{YeSpatialHandECCV2016}
Q.~Ye, S.~Yuan, T.-K. Kim, Spatial attention deep net with partial pso for
  hierarchical hybrid hand pose estimation, in: The European Conference on
  Computer Vision (ECCV), 2016, pp. 346--361.

\bibitem{Xu2017}
C.~Xu, L.~N. Govindarajan, Y.~Zhang, L.~Cheng, Lie-x: Depth image based
  articulated object pose estimation, tracking, and action recognition on lie
  groups, International Journal of Computer Vision (2017) 1--25\href
  {http://dx.doi.org/10.1007/s11263-017-0998-6}
  {\path{doi:10.1007/s11263-017-0998-6}}.

\bibitem{guo2017region}
H.~Guo, G.~Wang, X.~Chen, C.~Zhang, F.~Qiao, H.~Yang, Region ensemble network:
  Improving convolutional network for hand pose estimation, in: Image
  Processing (ICIP), 2017 IEEE International Conference on, IEEE, 2017.

\bibitem{wan2017crossing}
C.~Wan, T.~Probst, L.~Van~Gool, A.~Yao, Crossing nets: Dual generative models
  with a shared latent space for hand pose estimation, in: Proceedings of the
  IEEE Conference on Computer Vision and Pattern Recognition, 2017.

\bibitem{ge2017threedcnn}
L.~Ge, H.~Liang, J.~Yuan, D.~Thalmann, 3d convolutional neural networks for
  efficient and robust hand pose estimation from single depth images, in:
  Proceedings of the IEEE Conference on Computer Vision and Pattern
  Recognition, 2017, pp. 1991--2000.

\bibitem{zhang2012microsoft}
Z.~Zhang, Microsoft kinect sensor and its effect, IEEE multimedia 19~(2) (2012)
  4--10.

\bibitem{wang2013depth}
G.~Wang, X.~Yin, X.~Pei, C.~Shi, Depth estimation for speckle projection system
  using progressive reliable points growing matching, Applied optics 52~(3)
  (2013) 516--524.

\bibitem{shi2015high}
C.~Shi, G.~Wang, X.~Yin, X.~Pei, B.~He, X.~Lin, High-accuracy stereo matching
  based on adaptive ground control points, IEEE Transactions on Image
  Processing 24~(4) (2015) 1412--1423.

\bibitem{keselman2017intel}
L.~Keselman, J.~I. Woodfill, A.~Grunnet-Jepsen, A.~Bhowmik, Intel realsense
  stereoscopic depth cameras, arXiv preprint arXiv:1705.05548.

\bibitem{ge2016robust}
L.~Ge, H.~Liang, J.~Yuan, D.~Thalmann, Robust 3d hand pose estimation in single
  depth images: from single-view cnn to multi-view cnns, in: Proceedings of the
  IEEE Conference on Computer Vision and Pattern Recognition, 2016, pp.
  3593--3601.

\bibitem{oberweger2015hands}
M.~Oberweger, P.~Wohlhart, V.~Lepetit, Hands deep in deep learning for hand
  pose estimation, in: Proceedings of Computer Vision Winter Workshop, 2015,
  2015, pp. 21--30.

\bibitem{oberweger2015training}
M.~Oberweger, P.~Wohlhart, V.~Lepetit, Training a feedback loop for hand pose
  estimation, in: Proceedings of the IEEE International Conference on Computer
  Vision, 2015, pp. 3316--3324.

\bibitem{madadi2017end}
M.~Madadi, S.~Escalera, X.~Baro, J.~Gonzalez, End-to-end global to local cnn
  learning for hand pose recovery in depth data, arXiv preprint
  arXiv:1705.09606.

\bibitem{zhou2016model}
X.~Zhou, Q.~Wan, W.~Zhang, X.~Xue, Y.~Wei, Model-based deep hand pose
  estimation, in: Proceedings of the Twenty-Fifth International Joint
  Conference on Artificial Intelligence (IJCAI-16), 2016, pp. 2421--2427.

\bibitem{du2015hierarchical}
Y.~Du, W.~Wang, L.~Wang, Hierarchical recurrent neural network for skeleton
  based action recognition, in: Proceedings of the IEEE conference on computer
  vision and pattern recognition, 2015, pp. 1110--1118.

\bibitem{tang2014latent}
D.~Tang, H.~Jin~Chang, A.~Tejani, T.-K. Kim, Latent regression forest:
  Structured estimation of 3d articulated hand posture, in: Proceedings of the
  IEEE Conference on Computer Vision and Pattern Recognition, 2014, pp.
  3786--3793.

\bibitem{sun2015cascaded}
X.~Sun, Y.~Wei, S.~Liang, X.~Tang, J.~Sun, Cascaded hand pose regression, in:
  Proceedings of the IEEE Conference on Computer Vision and Pattern
  Recognition, 2015, pp. 824--832.

\bibitem{wan2016hand}
C.~Wan, A.~Yao, L.~Van~Gool, Hand pose estimation from local surface normals,
  in: European Conference on Computer Vision, Springer, 2016, pp. 554--569.

\bibitem{valentin2016learning}
J.~Valentin, A.~Dai, M.~Nie{\ss}ner, P.~Kohli, P.~Torr, S.~Izadi, C.~Keskin,
  Learning to navigate the energy landscape, in: 3D Vision (3DV), 2016 Fourth
  International Conference on, IEEE, 2016, pp. 323--332.

\bibitem{zhang2016learning}
Y.~Zhang, C.~Xu, L.~Cheng, Learning to search on manifolds for 3d pose
  estimation of articulated objects, arXiv preprint arXiv:1612.00596.

\bibitem{tagliasacchi2015robust}
A.~Tagliasacchi, M.~Schr{\"o}der, A.~Tkach, S.~Bouaziz, M.~Botsch, M.~Pauly,
  Robust articulated-icp for real-time hand tracking, in: Computer Graphics
  Forum, Vol.~34, Wiley Online Library, 2015, pp. 101--114.

\bibitem{tkach2016sphere}
A.~Tkach, M.~Pauly, A.~Tagliasacchi, Sphere-meshes for real-time hand modeling
  and tracking, ACM Transactions on Graphics (TOG) 35~(6) (2016) 222.

\bibitem{joseph2016fits}
D.~Joseph~Tan, T.~Cashman, J.~Taylor, A.~Fitzgibbon, D.~Tarlow, S.~Khamis,
  S.~Izadi, J.~Shotton, Fits like a glove: Rapid and reliable hand shape
  personalization, in: Proceedings of the IEEE Conference on Computer Vision
  and Pattern Recognition, 2016, pp. 5610--5619.

\bibitem{taylor2016efficient}
J.~Taylor, L.~Bordeaux, T.~Cashman, B.~Corish, C.~Keskin, T.~Sharp, E.~Soto,
  D.~Sweeney, J.~Valentin, B.~Luff, et~al., Efficient and precise interactive
  hand tracking through joint, continuous optimization of pose and
  correspondences, ACM Transactions on Graphics (TOG) 35~(4) (2016) 143.

\bibitem{krejov2015combining}
P.~Krejov, A.~Gilbert, R.~Bowden, Combining discriminative and model based
  approaches for hand pose estimation, in: Automatic Face and Gesture
  Recognition (FG), 2015 11th IEEE International Conference and Workshops on,
  Vol.~1, IEEE, 2015, pp. 1--7.

\bibitem{choi2015collaborative}
C.~Choi, A.~Sinha, J.~Hee~Choi, S.~Jang, K.~Ramani, A collaborative filtering
  approach to real-time hand pose estimation, in: Proceedings of the IEEE
  International Conference on Computer Vision, 2015, pp. 2336--2344.

\bibitem{sridhar2015fast}
S.~Sridhar, F.~Mueller, A.~Oulasvirta, C.~Theobalt, Fast and robust hand
  tracking using detection-guided optimization, in: Proceedings of the IEEE
  Conference on Computer Vision and Pattern Recognition, 2015, pp. 3213--3221.

\bibitem{sharp2015accurate}
T.~Sharp, C.~Keskin, D.~Robertson, J.~Taylor, J.~Shotton, D.~Kim, C.~Rhemann,
  I.~Leichter, A.~Vinnikov, Y.~Wei, et~al., Accurate, robust, and flexible
  real-time hand tracking, in: Proceedings of the 33rd Annual ACM Conference on
  Human Factors in Computing Systems, ACM, 2015, pp. 3633--3642.

\bibitem{qian2014realtime}
C.~Qian, X.~Sun, Y.~Wei, X.~Tang, J.~Sun, Realtime and robust hand tracking
  from depth, in: Proceedings of the IEEE Conference on Computer Vision and
  Pattern Recognition, 2014, pp. 1106--1113.

\bibitem{tang2013real}
D.~Tang, T.-H. Yu, T.-K. Kim, Real-time articulated hand pose estimation using
  semi-supervised transductive regression forests, in: Proceedings of the IEEE
  international conference on computer vision, 2013, pp. 3224--3231.

\bibitem{liang2014parsing}
H.~Liang, J.~Yuan, D.~Thalmann, Parsing the hand in depth images, IEEE
  Transactions on Multimedia 16~(5) (2014) 1241--1253.

\bibitem{zhu2015face}
S.~Zhu, C.~Li, C.~Change~Loy, X.~Tang, Face alignment by coarse-to-fine shape
  searching, in: Proceedings of the IEEE Conference on Computer Vision and
  Pattern Recognition, 2015, pp. 4998--5006.

\bibitem{chen2014joint}
D.~Chen, S.~Ren, Y.~Wei, X.~Cao, J.~Sun, Joint cascade face detection and
  alignment, in: European Conference on Computer Vision, Springer, 2014, pp.
  109--122.

\bibitem{kowalski2017deep}
M.~Kowalski, J.~Naruniec, T.~Trzcinski, Deep alignment network: A convolutional
  neural network for robust face alignment, arXiv preprint arXiv:1706.01789.

\bibitem{toshev2014deeppose}
A.~Toshev, C.~Szegedy, Deeppose: Human pose estimation via deep neural
  networks, in: Proceedings of the IEEE Conference on Computer Vision and
  Pattern Recognition, 2014, pp. 1653--1660.

\bibitem{carreira2016human}
J.~Carreira, P.~Agrawal, K.~Fragkiadaki, J.~Malik, Human pose estimation with
  iterative error feedback, in: Proceedings of the IEEE Conference on Computer
  Vision and Pattern Recognition, 2016, pp. 4733--4742.

\bibitem{maas2013rectifier}
A.~L. Maas, A.~Y. Hannun, A.~Y. Ng, Rectifier nonlinearities improve neural
  network acoustic models, in: in ICML Workshop on Deep Learning for Audio,
  Speech and Language Processing, Citeseer, 2013.

\bibitem{lin2000modeling}
J.~Lin, Y.~Wu, T.~S. Huang, Modeling the constraints of human hand motion, in:
  Human Motion, 2000. Proceedings. Workshop on, IEEE, 2000, pp. 121--126.

\bibitem{wu2001hand}
Y.~Wu, T.~S. Huang, Hand modeling, analysis and recognition, IEEE Signal
  Processing Magazine 18~(3) (2001) 51--60.

\bibitem{jia2014caffe}
Y.~Jia, E.~Shelhamer, J.~Donahue, S.~Karayev, J.~Long, R.~Girshick,
  S.~Guadarrama, T.~Darrell, Caffe: Convolutional architecture for fast feature
  embedding, arXiv preprint arXiv:1408.5093.

\bibitem{girshick2015fast}
R.~Girshick, Fast r-cnn, in: Proceedings of the IEEE international conference
  on computer vision, 2015, pp. 1440--1448.

\bibitem{melax2013dynamics}
S.~Melax, L.~Keselman, S.~Orsten, Dynamics based 3d skeletal hand tracking, in:
  Proceedings of Graphics Interface 2013, Canadian Information Processing
  Society, 2013, pp. 63--70.

\bibitem{he2017mask}
K.~He, G.~Gkioxari, P.~Doll{\'a}r, R.~Girshick, Mask r-cnn, in: Proceedings of
  the IEEE international conference on computer vision, 2017.

\bibitem{fourure2017multi}
D.~Fourure, R.~Emonet, E.~Fromont, D.~Muselet, N.~Neverova, A.~Tr{\'e}meau,
  C.~Wolf, Multi-task, multi-domain learning: application to semantic
  segmentation and pose regression, Neurocomputing 251 (2017) 68--80.

\bibitem{wang2018region}
G.~Wang, X.~Chen, H.~Guo, C.~Zhang, Region ensemble network: Towards good
  practices for deep 3d hand pose estimation, Journal of Visual Communication
  and Image Representation\href
  {http://dx.doi.org/https://doi.org/10.1016/j.jvcir.2018.04.005}
  {\path{doi:https://doi.org/10.1016/j.jvcir.2018.04.005}}.

\bibitem{madadi2017occlusion}
M.~Madadi, S.~Escalera, A.~Carruesco, C.~Andujar, X.~Bar{\'o}, J.~Gonz{\`a}lez,
  Occlusion aware hand pose recovery from sequences of depth images, in:
  Automatic Face \& Gesture Recognition (FG 2017), 2017 12th IEEE International
  Conference on, IEEE, 2017, pp. 230--237.

\bibitem{sinha2016deephand}
A.~Sinha, C.~Choi, K.~Ramani, Deephand: Robust hand pose estimation by
  completing a matrix imputed with deep features, in: Proceedings of the IEEE
  Conference on Computer Vision and Pattern Recognition, 2016, pp. 4150--4158.

\bibitem{ren2015faster}
S.~Ren, K.~He, R.~Girshick, J.~Sun, Faster r-cnn: Towards real-time object
  detection with region proposal networks, in: Advances in neural information
  processing systems, 2015, pp. 91--99.

\end{thebibliography}

\end{document}